\documentclass[journal]{IEEEtran}
\usepackage{graphicx}
\usepackage{multicol}
\usepackage{times}
\usepackage{epsfig}
\usepackage{graphicx}
\usepackage{amsmath}
\usepackage{amssymb}
\usepackage{algorithm,algorithmic}
\usepackage{algorithmic}
\usepackage{multirow}
\usepackage{color}
\usepackage[table ]{ xcolor}
\usepackage{booktabs}
\usepackage{stackengine}
\usepackage{hyperref}
\usepackage{array, makecell, tabularx, graphicx}
\ifCLASSINFOpdf
\else
\fi

\begin{document}
%
\title{CogRail: Benchmarking VLMs in \\ Cognitive Intrusion Perception for \\ Intelligent Railway Transportation Systems}
%
%
%

\author{Yonglin~Tian$^{1}$$^{*}$, Qiyao~Zhang$^{2}$$^{*}$, Wei~Xu$^{3}$, Yutong~Wang$^{1}$,  Yihao~Wu$^{4}$,  Xinyi~Li$^{4}$, Xingyuan~Dai$^{1}$, Hui~Zhang$^{5}$, Zhiyong~Cui$^{1}$, Baoqing~Guo$^{6}$, ~\IEEEmembership{Member,~IEEE}, Zujun~Yu$^{6}$, and Yisheng~Lv$^{1}$, ~\IEEEmembership{Member,~IEEE} 

\thanks{* These authors contributed equally to this work.}
\thanks{This work is partly supported by National Natural Science Foundation of China (62303460, 52441202), Beijing Natural Science Foundation-Fengtai Rail Transit Frontier Research Joint Fund (L231002), The Science and Technology Development Fund of Macau SAR (No. 0145/2023/RIA3 and
0093/2023/RIA2), and the Young Elite Scientists Sponsorship Program of China Association of Science and Technology under Grant YESS20220372. Corresponding author: Baoqing Guo, Yisheng Lv.}
\thanks{$^{1}$Yonglin Tian, Yutong Wang, Xingyuan Dai and Yisheng Lv are with the State Key Laboratory of Multimodal Artificial Intelligence Systems, Institute of Automation, Chinese Academy of Sciences, Beijing 100190, China.  {\tt\small yonglin.tian@ia.ac.cn, yutong.wang@ia.ac.cn, xingyuan.dai@ia.ac.cn, yisheng.lv@ia.ac.cn}}
\thanks{$^{2}$Qiyao Zhang is with the School of Automation, Beijing Institute of Technology, Beijing 100081, China. {\tt\small 3220241221@bit.edu.cn}}
\thanks{$^{3}$Wei Xu is with the Signal \& Communication Research Institute, China Academy of Railway Sciences. {\tt\small wei.xu@caa.org.cn}}
\thanks{$^{4}$Yihao Wu and Xinyi Li are with the internship of Beijing Huairou  Academy of Parallel Sensing, Beijing, China. {\tt\small 120213206080@stu.ustl.edu.cn, leexinyi2025@gmail.com}}
\thanks{$^{5}$Hui Zhang is with the School of Computer Science and Technology, Beijing Jiaotong University. {\tt\small huizhang1@bjtu.edu.cn}}
\thanks{$^{6}$Zhiyong Cui is with State Key Lab of Intelligent Transportation Systems, School of Transportation Science and Engineering, Beihang University. {\tt\small zhiyongc@buaa.edu.cn}}

\thanks{$^{7}$Baoqing Guo and Zujun Yu are with the State Key Laboratory of Advanced Rail Autonomous Operation, Beijing Jiaotong University, Beijing, 100044, China. {\tt\small bqguo@bjtu.edu.cn, zjyu@bjtu.edu.cn}}

}


\markboth{Journal of \LaTeX\ Class Files,~Vol.~14, No.~8, August~2015}%
{Shell \MakeLowercase{\textit{et al.}}: Bare Demo of IEEEtran.cls for IEEE Journals}

\maketitle
\begin{abstract}
Accurate and early perception of potential intrusion targets is essential for ensuring the safety of railway transportation systems. However, most existing systems focus narrowly on object classification within fixed visual scopes and apply rule-based heuristics to determine intrusion status, often overlooking targets that pose latent intrusion risks. Anticipating such risks requires the cognition of spatial context and temporal dynamics for the object of interest (OOI), which presents challenges for conventional visual models. To facilitate deep intrusion perception, we introduce a novel benchmark, CogRail, which integrates curated open-source datasets with cognitively driven question-answer annotations to support spatio-temporal reasoning and prediction. Building upon this benchmark, we conduct a systematic evaluation of state-of-the-art visual-language models (VLMs) using multimodal prompts to identify their strengths and limitations in this domain. Furthermore, we fine-tune VLMs for better performance and propose a joint fine-tuning framework that integrates three core tasks, position perception, movement prediction, and threat analysis, facilitating effective adaptation of general-purpose foundation models into specialized models tailored for cognitive intrusion perception. Extensive experiments reveal that current large-scale multimodal models struggle with the complex spatial-temporal reasoning required by the cognitive intrusion perception task, underscoring the limitations of existing foundation models in this safety-critical domain. In contrast, our proposed joint fine-tuning framework significantly enhances model performance by enabling targeted adaptation to domain-specific reasoning demands, highlighting the advantages of structured multi-task learning in improving both accuracy and interpretability. Code will be available at \href{https://github.com/Hub-Tian/CogRail}{https://github.com/Hub-Tian/CogRail}.
\end{abstract}

\begin{IEEEkeywords}
Intrusion perception, intelligent railway transportation systems, visual-language models, foundation models, AI agents.
\end{IEEEkeywords}

%
\IEEEpeerreviewmaketitle

\section{Introduction}

Intelligent intrusion perception plays a vital role in safeguarding the operational safety of railway transportation systems \cite{cao2024railway, SUN2025127005}. In recent years, the expansion of railway routes and the continuous increase in train speeds \cite{xu2023transformer} have significantly heightened the risk posed by intrusions into the railway perimeter. Incursions by pedestrians, animals, or vehicles have become increasingly prominent safety concerns, with the potential to cause severe disruptions or accidents. These trends impose greater demands on perception systems of the object of interest (OOI) \cite{chen2025real}, not only for timely and accurate detection but also for predictive awareness of possible intrusions. Addressing these challenges is critical for the development of intelligent and reliable railway transportation systems \cite{li2023research}.

Recent studies have explored a variety of methods for railway intrusion detection, ranging from contact-based sensors to non-contact approaches driven by machine learning and deep learning techniques \cite{wang2024review}. Contact-based methods typically focus on determining the presence or position of objects through physical interaction with the environment \cite{catalanoOpticalFiberIntrusion2017}, while non-contact approaches, such as vision-based or learning-based systems \cite{lian2024continuous}, emphasize the classification, detection, or segmentation of targets based on semantic analysis \cite{7015584}. Although these approaches enable a certain level of intrusion awareness, they suffer from notable limitations. Contact-based systems often fail to distinguish between different categories of intrusion targets, limiting their ability to support context-aware responses. On the other hand, non-contact approaches tend to face challenges in robustness and generalization, particularly in complex or unseen scenarios. Moreover, both paradigms generally exhibit limited capacity for anticipatory reasoning, making it difficult to perceive objects with latent intrusion tendencies.

To better address the limitations of existing intrusion detection approaches, there is a growing need to develop and formalize the concept of cognitive railway intrusion perception. Instead of merely identifying objects that have already breached safety boundaries, cognitive perception can achieve the early recognition and reasoning of entities that may exhibit potential intrusion intent, even before any explicit violations occur. This shift from reactive detection to anticipatory perception highlights the importance of proactive safety mechanisms, particularly in complex and dynamic railway environments.

Cognitive intrusion perception involves three interrelated dimensions: semantic awareness, positional awareness, and motion awareness. Semantic awareness concerns the categorization of objects by type; positional awareness refers to the analysis of spatial relationships between objects and railway infrastructure; and motion awareness involves modeling the temporal dynamics of object movement and their potential transition toward intrusion. Among these dimensions, semantic awareness has received considerable attention in recent years, supported by the emergence of various annotated datasets (as shown in Table \ref{Existing image datasets}) and the rapid progress of deep learning techniques for object detection and classification. In contrast, positional awareness—particularly in relation to specific components of the railway infrastructure such as tracks, ballast, and danger zones—still largely relies on heuristic rules or contact-based sensors. Purely vision-based approaches for modeling such positional relationships remain underdeveloped due to the scarcity of structured data and the lack of task-specific methodologies. Similarly, motion awareness, which emphasizes understanding the relative movement of objects with respect to the railway, is still in its infancy. 


\begin{table*}[htbp]
\centering
\caption{Existing datasets for track intrusion detection}
\label{Existing image datasets}
\renewcommand{\arraystretch}{1.2}
\setlength{\tabcolsep}{3pt} 
\footnotesize 
\begin{tabularx}{\textwidth}{@{}p{2.2cm}p{3cm}p{5cm}p{5cm}p{1cm}p{1cm}@{}}
\toprule
Dataset                                                                                 & Data Type                                                 & Target Type                                                                                                                  & Annotation Type                                                                                                            & Open   & Year \\ \midrule
RailSem19 \cite{zendel2019railsem19}                                                                                      &RGB images                & Trains,   pedestrians, vehicles, etc. & Semantic   segmentation                                                           & Yes                    & 2019 \\
RAWPED \cite{toprakConditionalWeightedEnsemble2020}                                                &  RGB images                   & Pedestrians                                                     & Bounding boxes                                                                                               & No                     & 2020 \\
Literature \cite{MultiblockSSDBased2020}                                                       & RGB images                              & Pedestrians,   trains, stones                                        & Bounding boxes                                                   & No                     & 2020 \\
MRSI \cite{chen2022mrsi}                                                                                       & RGB, infrared images                       & Pedestrians,   bicycles, cars, track, etc.                                                           & Bounding boxes, semantic segmentation                & Yes                    & 2021 \\
Rail \cite{9380436} & RGB images, point clouds & Track &Semantic segmentation&  No & 2021 \\ 
Literature \cite{gongEnhancedFewShotLearning2022}                              & RGB images                                 & Pedestrians,   animals                                                                                   & Bounding boxes                                        & No                     & 2022 \\
Metro \cite{wangLRENetVisionBasedRealtime2023}                                         & RGB images                        & Track                                                                                                                   & Grid coordinate   segmentation                                         & No                     & 2023 \\
Literature \cite{huangRailwayIntrusionDetection2023} & RGB images                                  & Track, pedestrians, sundries                                                                                          & Bounding boxes                                                                                            & No                     & 2023 \\
Article \cite{liRailwayIntrusionDetection2023}                                                & RGB images                        & Pedestrians,   tracks, plastics, etc.                                                                                & Bounding boxes, semantic segmentation                                        & No                     & 2023 \\
RailAnomaly \cite{wangIntrusionDetectionHighspeed2023}                                                 & RGB images                       & Pedestrians                                                                                                                  & Classification labels & No                     & 2023 \\
RS \cite{mengSDRCYOLONovelForeign2023a}                                                           & RGB images   & Pedestrians, cars, bicycles                                            & Bounding boxes                                                                                             & No                     & 2023 \\
FSRT2023 \cite{yeFewShotRailwayIntrusion2024}                                                   & RGB images & Pedestrians,   tracks,  toolboxes, etc.                                                           &    Bounding boxes                                                                                               & No                     & 2024 \\ 
RailPSG\cite{10988713} & RGB images & Pedestrians, train, tracks, etc.& Semantic segmentation, relationship & No & 2025\\

\bottomrule
\end{tabularx}
\end{table*}

To promote the research and development of cognitive intrusion perception and to provide robust data and methodological foundations, we introduce the multimodal question answering dataset, CogRail, specifically designed for this task. Built upon open-source visual surveillance data, CogRail includes structured visual questions targeting two underexplored yet critical dimensions: relative positional awareness and relative motion awareness. In addition, we provide expert-annotated threat level assessments, enabling fine-grained reasoning about the intrusion risk of observed entities. Based on this dataset, we establish RailGPT, the first evaluation and fine-tuning framework for assessing and adapting mainstream vision-language foundation models in the railway safety domain. To further enhance model adaptability, we propose a multi-task joint tuning strategy that aligns model outputs across semantic understanding, spatial reasoning, and risk assessment subtasks. Our main contributions are summarized as follows.

\begin{itemize}
    \item We construct the first benchmark for cognitive railway intrusion perception, integrating multimodal visual question answering with expert-defined threat annotations grounded in real-world railway scenarios.
    
    \item This work introduces RailGPT, a unified perception architecture integrating mainstream open-source VLMs. We systematically evaluate the performance boundaries and domain gaps of SOTA VLMs and, through targeted fine-tuning, show enhanced perception, establishing a scalable foundation for railway intrusion perception systems.

    \item A multi-task fine-tuning framework for cognitive intrusion reasoning is introduced, successfully transferring multiple capabilities to open-source VLMs and yielding significant gains in perception accuracy across tasks.
\end{itemize}

The remainder of this paper is organized as follows. Section~\ref{sec:relatedwork} reviews related work on railway intrusion detection and vision-language models. Section~\ref{sec:dataset} presents the construction of our proposed dataset, \textit{CogRail}, including its annotation principles and task design. Section~\ref{sec:method} introduces our cognitive intrusion reasoning framework, including visual-textual prompting and the multi-task fine-tuning architecture. Section~\ref{sec:experiment} details the experimental setup, evaluation protocols, and results. Finally, Section~\ref{sec:conclusion} concludes the paper and outlines directions for future research.

\section{Related Work}
\label{sec:relatedwork}

\subsection{Railway Intrusion Detection}

Railway transportation is a highly safety-critical domain, where security concerns have a direct impact on both operational efficiency and overall system reliability \cite{zhou2025railway}. Among the various safety measures, intrusion detection serves as a key component, playing a crucial role in protecting infrastructure and preventing potential disruptions to railway operations. Existing methods on railway intrusion detection mainly fall into three categories: contact-based, traditional-vision-based, and deep-learning–based approaches. However, challenges remain in terms of scenario adaptability, semantic understanding of intrusions, and the scarcity of high-quality datasets \cite{wang2024review, shi2024survey}.


Traditional methods: 
Early studies employed physical sensing techniques, such as fiber Bragg gratings \cite{catalanoOpticalFiberIntrusion2017} and distributed acoustic sensing combined with neural networks \cite{liFiberDistributedAcoustic2020}. These systems offer strong immunity to electromagnetic interference; however, they lack fine-grained semantic understanding and exhibit limited generalization capabilities in dynamic or open environments.

Techniques based on image differencing \cite{7015584} and optical flow analysis \cite{6614039} have demonstrated moderate real-time performance but are prone to high false alarm rates and exhibit poor robustness in the presence of noise. Alternative strategies, including background modeling \cite{caoEffectiveRailwayIntrusion2022} and predefined danger-zone mapping \cite{wangAdaptiveTrackSegmentation2019}, have also been explored. Nevertheless, these methods remain sensitive to environmental variations such as changes in lighting conditions or vegetation, which frequently lead to false positives.

Deep-learning-based methods: With the advent of convolutional neural networks, railway intrusion detection has seen notable performance improvements. However, most existing models remain limited to object detection tasks and fail to incorporate intrusion-specific semantics. For example, YOLOv5s-based models can detect common intruders on railway tracks \cite{zhangRailwayObstacleIntrusion2024}, yet they are unable to assess intrusion risk levels based on motion patterns or spatial context. A major bottleneck in advancing these models lies in the limited availability of high-quality datasets. Table \ref{Existing image datasets} summarizes publicly available datasets related to railway visual perception, particularly for track intrusion detection tasks. Most existing datasets consist primarily of RGB images, reflecting the dominance of vision-based approaches in this domain. A few datasets extend beyond visible-spectrum imaging, incorporating infrared images (e.g., MRSI \cite{chen2022mrsi}) or 3D point cloud data (e.g., Rail \cite{9380436}), which are useful for enhancing robustness under poor lighting or complex geometric conditions. While these datasets offer valuable training resources, the majority are not publicly accessible. To ensure reproducibility and comprehensive coverage, our benchmark is constructed based on the only two publicly available datasets to date, RailSem19 \cite{zendel2019railsem19} and MRSI \cite{chen2022mrsi}, to the best of our knowledge.

Recently, some studies have attempted to track area segmentation \cite{wangLRENetVisionBasedRealtime2023, SUN2024109295}, ROI-based intrusion estimation \cite{wangAdaptiveTrackSegmentation2019}, or rule-based alarm triggering \cite{zhangRailwayObstacleIntrusion2024}, yet these methods often neglect temporal motion trends or proximity reasoning. Fusion-based frameworks integrating LiDAR \cite{9380436} or few-shot learning \cite{yeFewShotRailwayIntrusion2024} show promise, but remain limited by fixed thresholds or rigid rules.

\begin{figure*}[htbp]
\centering
\includegraphics[width=6.8in]{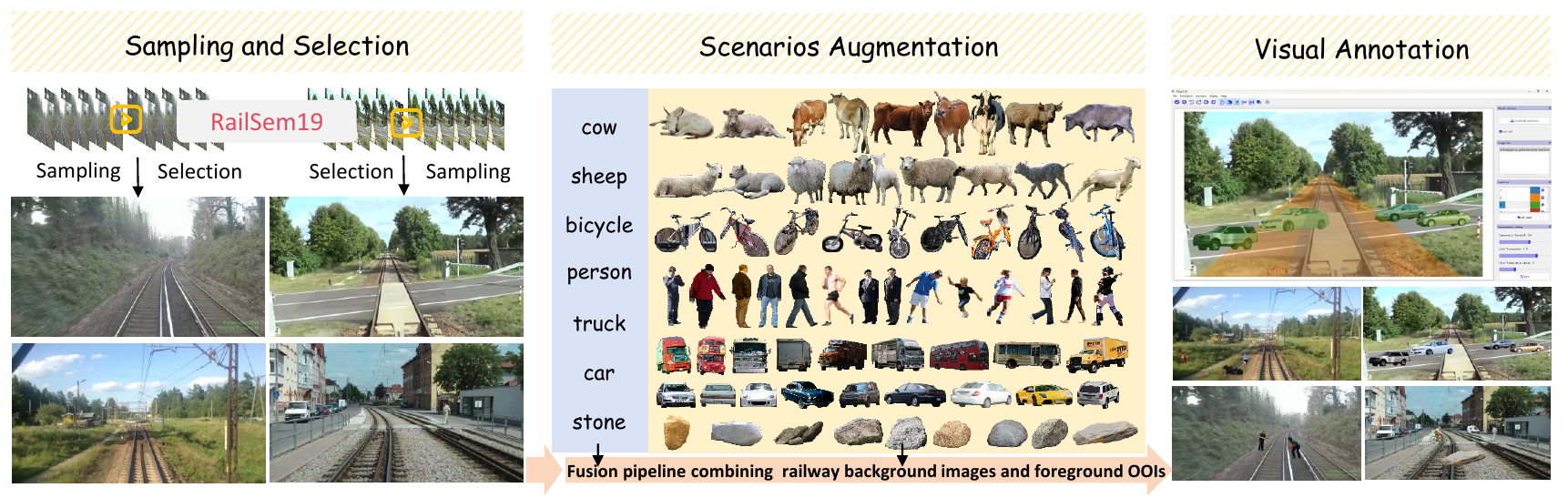}
\caption{Basic dataset construction pipeline on RailSem19}
\label{pipeline}
\end{figure*}

\begin{figure*}[htbp]
\centering
\includegraphics[width=6.8in]{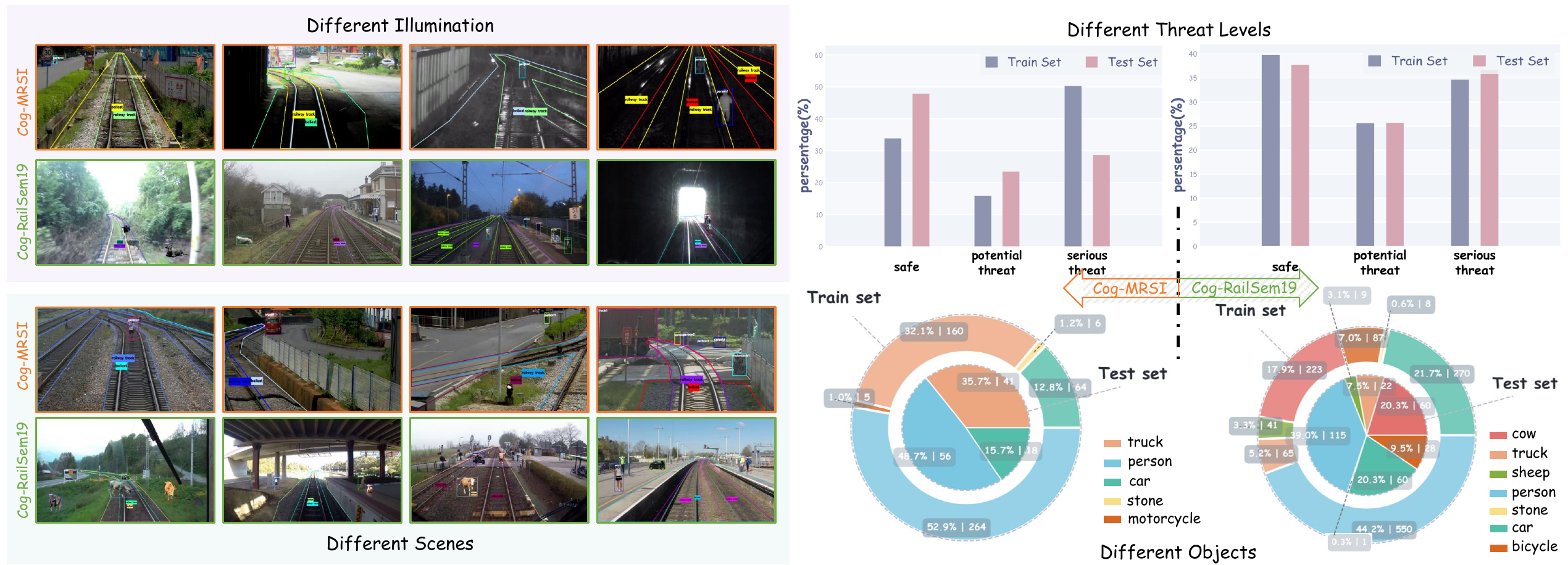}
\caption{Statistics of the proposed CogRail dataset}
\label{elements}
\end{figure*}

\subsection{Spatial Perception with Vision-Language Models}

Spatial reasoning is pivotal in safety-critical domains such as railway surveillance, autonomous driving, and robotics, where understanding object positions, motion trajectories, and their relationships with infrastructure directly impacts operational security.  VLMs contribute by fusing visual and linguistic cues to enhance understanding of spatial contexts, aiding in 3D scene cognition tasks like spatial reasoning \cite{wu2024prospective}. Traditional frameworks rely on structured 3D data, such as point clouds \cite{ye20223d, hong20233d}, but these require precise sensor calibration and are difficult to deploy in large-scale surveillance scenarios. 
To overcome such limitations, multimodal techniques such as LiDAR-LLM \cite{yang2025lidar} and LAGOR \cite{thomason2022language} enhance spatial localization by fusing LiDAR or multi-view image data. However, these approaches involve high computational costs and hardware dependencies, which limit their applicability in lightweight monitoring systems like trains or roadside cameras. In contrast, monocular-based VLMs such as SpatialVLM \cite{chen2024spatialvlm} and SpatialRGPT \cite{cheng2024spatialrgpt} have shown impressive spatial reasoning capabilities using only 2D inputs. These models estimate pseudo-depth, fuse geometric cues, and leverage large-scale pretraining (e.g., 2B QA pairs in SpatialVLM). Although efficient, they are highly data-hungry.

Another promising line of work is Spatial-MLLM \cite{wu2025spatial}, which uses 2D videos and implicit 3D representation learning to infer spatial relations. While it performs well even without depth inputs, such models still rely heavily on vast annotated datasets \cite{zhan2025open3dvqa, yang2024thinking}. Despite these efforts, three challenges persist: baseline spatial accuracy remains subpar compared to human perception \cite{liu2023visual}, multimodal methods face hardware and cost barriers \cite{yang2025lidar}, and data-intensive training limits real-world adaptability. In light of these limitations, this study focuses on a monocular image input paradigm and explores how large models’ contextual learning capabilities can support accurate spatial reasoning for railway intrusion detection.

\section{CogRail Benchmark}
\label{sec:dataset}

To promote the reasoning ability of intrusion perception models for intelligent railways, we propose a new benchmark named CogRail with the aim of addressing the dual challenges faced in the field of railway intrusion detection. Firstly, existing datasets such as MRSI \cite{chen2022mrsi} and RailSem19 \cite{zendel2019railsem19}  primarily focus on object classification and positional annotations, lacking critical descriptions of motion trends and scene semantics necessary for high-level tasks like potential threat prediction. Secondly, there is a significant shortage of visual-language interaction datasets suitable for multimodal foundation models, as traditional labeling paradigms do not meet the training requirements for spatiotemporal reasoning and causal logic.

To address these gaps, we constructed the CogRail benchmark dataset, which integrates spatial location, dynamic behavior, and threat semantics into a standardized evaluation framework for cognitive reasoning in railway intrusion detection.

\subsection{Task Definitions}
The CogRail benchmark includes three tasks that evaluate spatial perception, motion prediction, and threat assessment respectively in railway environments.

\subsubsection{RailPos: Spatial position perception}
This task infers the relative position of each object of interest (OOI) with respect to railway infrastructure, assigning one of three mutually exclusive labels: \textit{outside the railway}, where the OOI lies in safe areas beyond both track and ballast; \textit{on the ballast}, where the OOI is situated within the gravel bed adjacent to the track; and \textit{on the track}, where the OOI is directly on the rail. Accurate classification requires the model to internalize the spatial topology of railway elements, thereby providing a reliable spatial reference for downstream threat assessment.

\subsubsection{RailMove: Movement state prediction}
This task assesses how target motion patterns relate to railway safety by assigning one of three states: \textit{stationary}, for motionless OOI (e.g., standing pedestrians, parked vehicles); \textit{non‑threatening movement}, for OOI whose trajectories do not intersect the railway (e.g., vehicles traveling on a road); and \textit{threatening movement}, for targets whose current or forecast motion indicates potential intrusion into the track area (e.g., pedestrians moving toward the rails). The core challenge is to fuse temporal dynamics with spatial trajectories to enable predictive, risk‑aware reasoning.

\subsubsection{RailThreat: Threat level analysis}
This task synthesizes spatial positioning, motion state, and scene semantics to categorize threats into three levels: \textit{no threat}, indicating no potential for intrusion (e.g., animals in a distant field); \textit{potential threat}, denoting a latent risk of intrusion (e.g., pedestrians lingering on the ballast); and \textit{serious threat}, representing an existing threat (e.g., obstacles on the track). To align with real‑world safety warning needs, the model must integrate multi‑dimensional cues and reason jointly over space, time, and semantics to support reliable decision‑making.

\subsection{Dataset construction pipeline}
The dataset construction process consists of two main stages. The first is basic dataset construction, exemplified by RailSem19, where Figure \ref{pipeline} illustrates the workflow. This includes sampling and selecting railway background images, performing scenario augmentation to fuse foreground objects with background scenes, and conducting visual annotation for targets and regions. Building on this foundation, the second stage converts the visual annotations into instruction–response pairs enriched with linguistically diverse prompts, yielding a unified dataset format suitable for both perception-oriented tasks and instruction-tuned railway agents.

\subsubsection{Basic dataset construction}
{Backgrounds Sampling:} CogRail is constructed from the MRSI \cite{chen2022mrsi} and RailSem19 \cite{zendel2019railsem19} datasets. MRSI offers dense video frames with high inter-frame similarity, while RailSem19 provides a broader variety of scene types but contains relatively sparse foreground targets. From these sources, we select frames that include both railway tracks and relevant foreground objects, while ensuring a high degree of scene diversity and semantic variation across samples to promote robust generalization.

{Scenario augmentation:} To address the foreground sparsity in RailSem19, over 2,000 object instances, including cows, trucks, sheep, persons, stone, cars, and bicycles, are extracted from the LVIS dataset \cite{gupta2019lvis}. These instances are geometrically transformed in terms of scale, orientation, and spatial placement, guided by scene semantics and depth cues, and composited into the background images to enhance dataset diversity.

{Visual annotation:} Utilizing the EISeg tool \cite{hao2022eiseg}, tracks and ballast are annotated with polygonal segmentation, while other foreground objects (e.g., person, car, bicycle, animal, stones) use axis‑aligned bounding boxes with instance IDs. Each OOI additionally receives a semantic class, a spatial position tag, a motion state, and an overall threat level, yielding compact, task‑aligned supervision.

\subsubsection{Instruction-level dataset construction}
Building on the three sub‑tasks defined earlier, we convert the visual annotations and OOI category labels into instruction–response pairs. We first handcraft a compact set of seed dialogue templates for spatial position, motion state, and threat level. These seeds are then automatically expanded with GPT‑4 via paraphrasing, while enforcing consistency with the ground‑truth attributes. The resulting corpus is serialized as a JSON dataset for instruction tuning, with each record containing image identifiers, natural‑language instructions, optional context, and labels.

Our benchmark contains two subsets with harmonized labels (see Figure \ref{elements}). Cog‑MRSI includes 347 images with 614 OOIs across five classes (person, car, stone, truck, motorcycle) under a 4:1 train/test split; persons and trucks dominate, while stone and motorcycles are sparse. Its threat distribution is skewed toward higher risk in training (more Serious Impact) and safer cases in testing (more Safe), with Potential Impact in between. Cog‑RailSem19 comprises 495 images with 1,536 OOIs across seven classes (person, car, truck, bicycle, stone, cow, sheep). Animal and vehicle instances are extracted from LVIS instance masks \cite{gupta2019lvis} and manually composited into RailSem19 scenes by cut‑and‑paste, placing each at appropriate locations, orientations, and scales; their threat proportions are relatively balanced between train and test across Safe, Potential Impact, and Serious Impact. 

\begin{figure*}[htbp]
\centering
\includegraphics[width=6.8in]{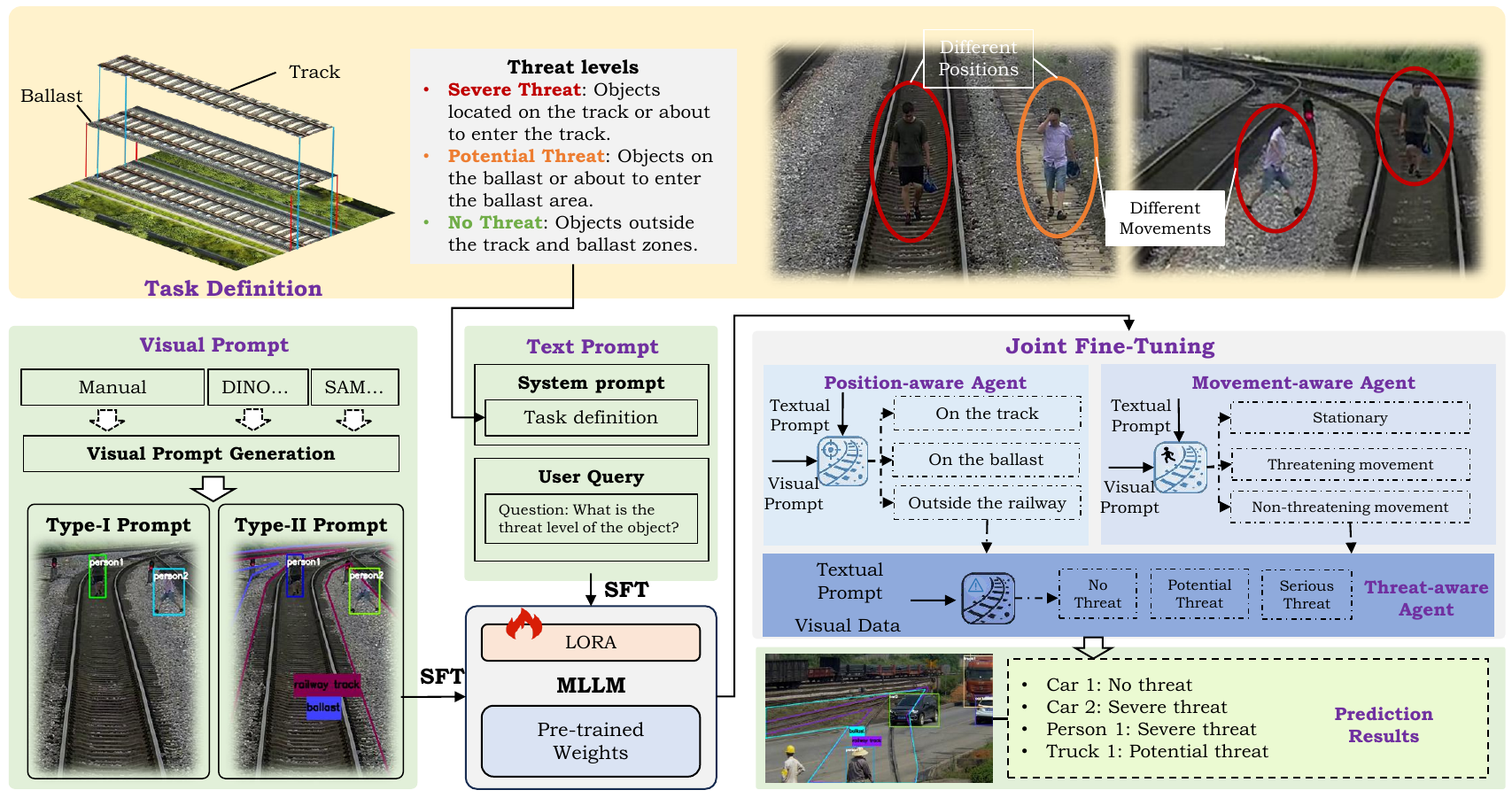}
\caption{The framework of the proposed RailGPT method.}
\label{framework}
\end{figure*}

\section{RAILGPT}
\label{sec:method} 
To address the cognitive reasoning challenges in railway intrusion perception, we propose RAILGPT (illustrated in Figure~\ref{framework}), a multimodal agent-based framework compatible with diverse types of VLMs. It comprises three components: multimodal prompting to ground perception and semantics, agents construction tailored to spatial positioning, motion, and threat assessment, and fine‑tuning of these agents to improve task-specific performance.

\subsection{Multimodal Prompting}

\subsubsection{Visual prompts}
Instead of focusing on small-scale visual models, such as general-purpose detectors (e.g., SAM \cite{zhang2022dino}, DINO \cite{kirillov2023segment}) or domain-specific perception models \cite{zhangRailwayObstacleIntrusion2024} optimized for railway scenes, RailGPT emphasizes the analysis of cognitive understanding in large multimodal models, particularly their ability to reason about potential intrusion scenarios based on perceptual inputs.

To guide large models in interpreting scene semantics, we design visual prompts based on the outputs of existing visual detectors. These prompts serve as structured, spatially grounded inputs that help foundation models focus on regions critical to intrusion perception. Specifically, we introduce two complementary types of visual hits. 

Object-level hits: which highlight dynamic, intrusion-prone targets such as pedestrians, animals, and vehicles. These are annotated using bounding boxes, following standard object detection conventions. This format not only facilitates alignment with pretrained visual encoders but also ensures compatibility with downstream reasoning tasks that require explicit object references.
     
Area-level hits: which define static components of the railway infrastructure, including tracks, ballast zones, and surrounding buffer areas. These regions are annotated using contour-based segmentation masks, allowing for precise delineation of operational boundaries and spatial zones relevant to safety reasoning.

Each visual hit is also assigned a unique index that is referenced in the corresponding textual prompt. This indexed design supports object-specific or region-specific reasoning and allows for fine-grained queries, such as assessing the motion state or threat level of a particular entity, thereby enhancing the interpretability and controllability of the multimodal model’s inference process.

Building upon these two categories of visual hits, we have developed two corresponding types of visual prompts: Type-I and Type-II. Specifically, Type-I visual prompts exclusively leverage object-level hits to explicitly indicate relevant objects, while Type-II visual prompts integrate both levels of hits concurrently to jointly highlight both the target objects and railway zones.

\subsubsection{Textual prompts}
The core of textual prompting is a structured system prompt that integrates four elements. First, a task context provides a concise overview of the railway intrusion detection objectives, establishing a foundational understanding. Second, a task description gives explicit instructions for determining the spatial position, motion state, or threat level of a target. Third, judge criteria define the benchmark-based rules that ensure consistent interpretation across scenarios. Finally, an output format specifies a structured response representation, improving interpretability and enabling reliable downstream processing.

\begin{figure}[htbp]
\centering
\includegraphics[width=3.3in]{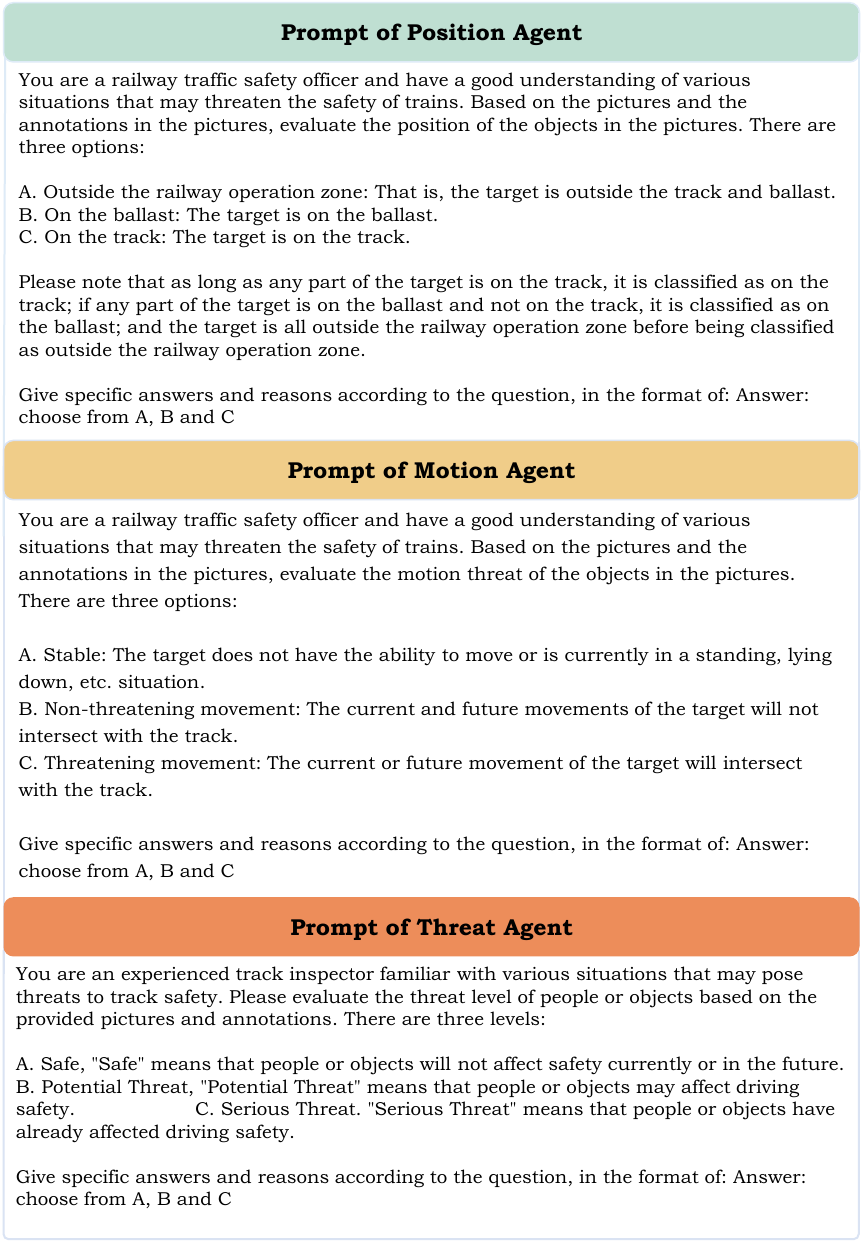}
\caption{The prompts for different agents.}
\label{elements-1}
\end{figure}

Figure \ref{elements-1} illustrates the prompts for position-aware agent, motion-aware agent and threat-aware agent. This design enables large language models to rapidly align with domain-specific goals by grounding their reasoning in pretraining knowledge and domain-defined constraints.

\subsection{Railway agents configuration}

To address the three core tasks defined in Section \ref{sec:dataset}, i.e., spatial positioning reasoning (RailPos), motion state classification (RailMove), and threat level assessment (RailThreat), we design specialized cognitive agents, each tailored to a distinct reasoning dimension through prompt engineering. The agent architecture is rooted in the CogRail benchmark’s task definitions, with each agent’s prompt meticulously crafted to align with the task descriptions and judgment criteria outlined in Figure \ref{elements}. 

The position-aware agent focuses on RailPos, tasked with inferring the relative spatial relationship between targets and railway infrastructure (e.g., on-track, on-ballast, or outside operational zones). The movement-aware agent targets RailMove, analyzing scene information and target state to classify motion states (stationary, non-threatening, or threatening). The intrusion analysis agent addresses RailThreat, synthesizing spatial positioning, motion states, and other signals to assess threat levels (safe, potential impact, serious impact).

Each agent’s prompt structure follows the standardized framework: task context, detailed instructions, judgment rules, and structured output formats, as defined in the CogRail benchmark. This design ensures modularity and interpretability, enabling the agents to collaboratively reason about intrusion scenarios while maintaining focus on their respective tasks.

\subsection{Fine-tuning of railway agents}
\subsubsection{Individual Fine-tuning}

To facilitate effective domain adaptation, we employ a structured approach to dataset construction and model tuning, ensuring robustness and generalization in railway intrusion perception. The training and testing datasets are formulated as visual-question-answer pairs, where each sample consists of a visual input and a textual query about the target’s positional relationship, motion state, or threat level. 

To mitigate the risk of model overfitting due to repetitive query patterns, we leverage a pre-trained large language model to generate semantically equivalent paraphrases of each query. This strategy diversifies the input space, encouraging the model to learn abstract reasoning rather than memorizing specific question formats. For instance, a query like "What is the movement state of person 1?" might be rewritten as "Of the three movement states, which one applies to person 1?" This process ensures that the model adapts to varied linguistic expressions while maintaining annotation consistency.

Given the inherent imbalance in object attribute distributions (e.g., serious impact threats are naturally less frequent than safe scenarios), we implement a proportional sampling strategy during dataset generation. Let $p_c$ denote the empirical proportion of class $c$ in the dataset, and $\alpha_c$ be the target sampling ratio. The resampling probability of class $c$ is defined as:
\begin{equation}
    P(c) = \frac{\alpha_c}{p_c} \Big/ \sum_{c'} \frac{\alpha_{c'}}{p_{c'}}.
\end{equation}

To further adapt the model to railway-specific scenarios, we employ Low-Rank Adaptation (LoRA) \cite{hu2022lora} for supervised fine-tuning. For supervised fine-tuning with LoRA, the updated parameterization becomes:
\begin{equation}
    W' = W + \Delta W, \quad \Delta W = BA,
\end{equation}
and the optimization objective is
\begin{equation}
    \theta^* = \arg\min_{\theta} \; \mathbb{E}_{(x, y) \sim \mathcal{D}} \big[ \mathcal{L}(f_{\theta}(x), y) \big],
\end{equation}
where $\theta = \{A, B\}$ are the learnable low-rank matrices, and $f_\theta$ is the adapted model. This technique modifies only a small portion of model parameters, significantly reducing training overhead.

The optimization objective is
\begin{equation}
    \theta^* = \arg\min_{\theta} \; \mathbb{E}_{(x, y) \sim \mathcal{D}} \big[ \mathcal{L}(f_{\theta}(x), y) \big],
\end{equation}
where $\theta = \{A, B\}$ are the learnable low-rank matrices, and $f_\theta$ is the adapted model. The task-specific cross-entropy loss is defined as:
\begin{equation}
\mathcal{L}_{t} = - \frac{1}{n_t} \sum_{i=1}^{n_t} \sum_{c=1}^{C_t} y^{t}_{i,c} \log \hat{y}^{t}_{i,c},
\end{equation}
where $C_t$ is the number of classes in task $t$, and $y^{t}_i$ and $\hat{y}^{t}_i$ denote the ground-truth label and predicted probability for task 
$t \in \{\mathrm{RailPos}, \mathrm{RailMove}, \mathrm{RailThreat}\}$, and $n_t$ is the number of samples in task $t$.

\subsubsection{Joint Fine-tuning}

Given the inherent interconnections among spatial positioning, motion state, and threat assessment, we further implement joint training by merging the three task datasets. By mixing three kinds of question-answer pairs in training dataset, model could learn to propagate dependencies across tasks.

The training data incorporates triplets consisting of positional labels (\textit{RailPos}), motion state (\textit{RailMove}), and threat level (\textit{RailThreat}), enabling multi-task optimization. The overall loss is then defined as follows.
\begin{equation}
\mathcal{L} = \lambda_{\mathrm{pos}}\mathcal{L}_{\mathrm{pos}} 
+ \lambda_{\mathrm{move}}\mathcal{L}_{\mathrm{move}} 
+ \lambda_{\mathrm{threat}}\mathcal{L}_{\mathrm{threat}}.
\end{equation}

\section{Experiments}
\label{sec:experiment}

\begin{figure*}[htbp]
\centering
\includegraphics[width=7in]{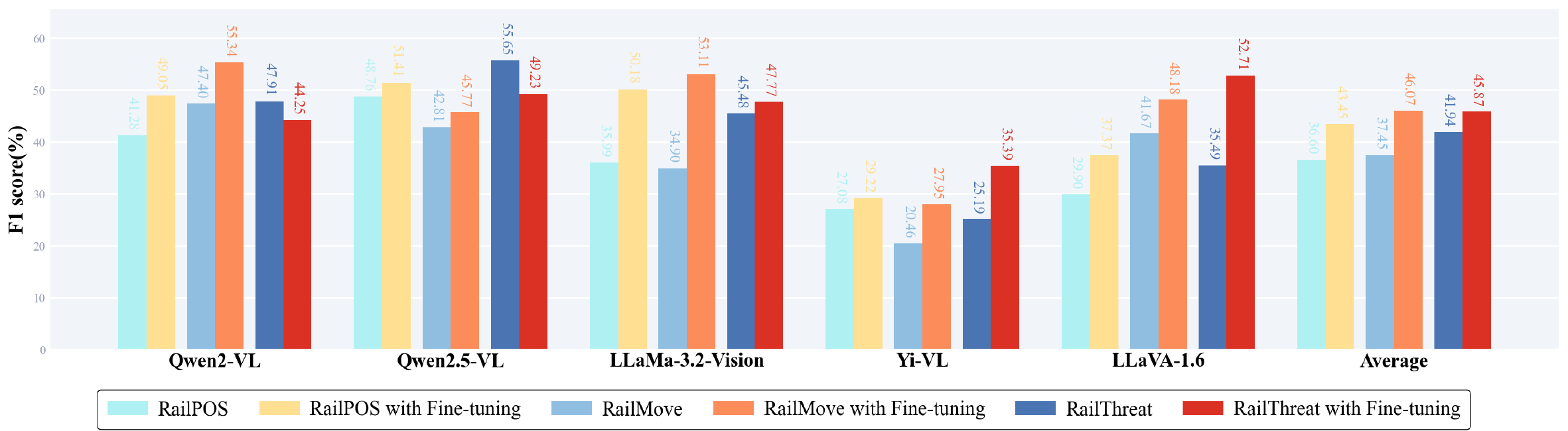}
\caption{Performance(F1) Comparison on Type-I Visual Prompt in Cog-MRSI dataset via Individual Fine-tuning}
\label{ex_data_show_1}
\end{figure*}

\begin{figure*}[htbp]
\centering
\includegraphics[width=7in]{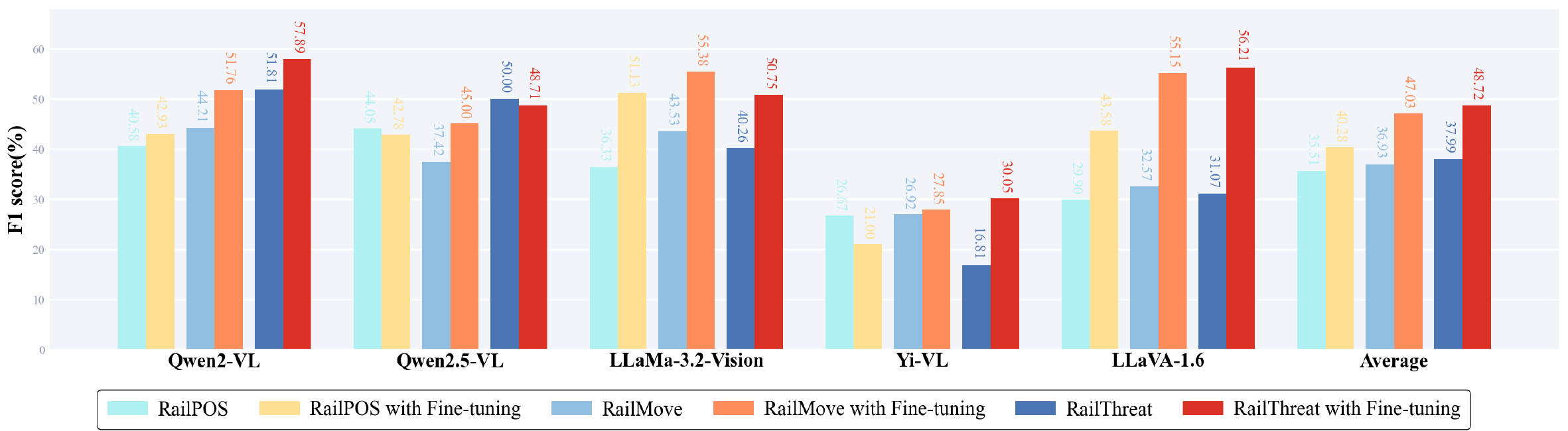}
\caption{Performance(F1) Comparison on Type-II Visual Prompt in Cog-MRSI dataset via Individual Fine-tuning}
\label{ex_data_show_2}
\end{figure*}

\begin{figure*}[htbp]
\centering
\includegraphics[width=7in]{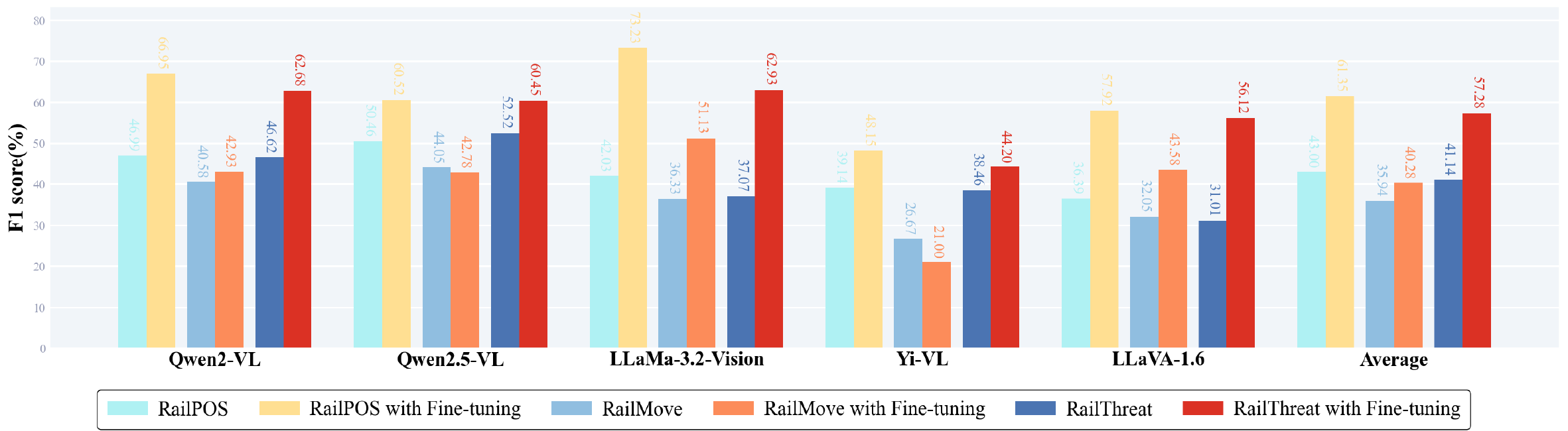}
\caption{Performance(F1) Comparison on Type-I Visual Prompt in Cog-RailSem19 dataset via Individual Fine-tuning}
\label{ex_data_show_3}
\end{figure*}

\begin{figure*}[htbp]
\centering
\includegraphics[width=7in]{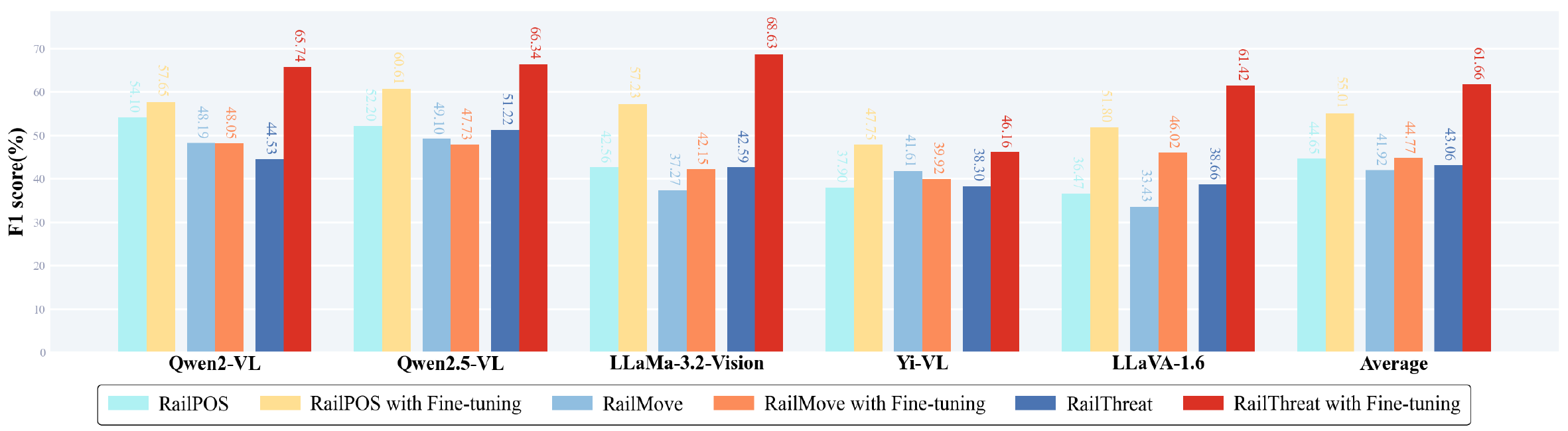}
\caption{Performance (F1) Comparison on Type-II Visual Prompt in Cog-RailSem19 dataset via Individual Fine-tuning}
\label{ex_data_show_4}
\end{figure*}

\begin{table*}[htbp]
\centering
\caption{Performance of Joint Fine-tuning with Type-I Visual Prompt for RailThreat Task on Cog-MRSI Dataset}
\label{tab:sota_mllm_type1-threat-MRSI-joint}
\renewcommand{\arraystretch}{1.3}
\resizebox{\textwidth}{!}{
\begin{tabular}{ccccccccccc}
\toprule
\multirow{2}{*}{\textbf{Model}} & \multirow{2}{*}{\textbf{Fine-tuned}} & \multicolumn{2}{c}{\textbf{Safe}} & \multicolumn{2}{c}{\textbf{Potential Threat}} & \multicolumn{2}{c}{\textbf{Serious Threat}} & \multicolumn{3}{c}{\textbf{Overall}} \\
\cline{3-11}
 & & Precision & Recall & Precision & Recall & Precision & Recall & Precision & Recall & \textbf{F1 Score} \\
\midrule
\multirow{3}{*}{Qwen2-VL} & No & 53.19 & 67.57 & 45.16 & 58.33 & 100.00 & 32.00 & 66.12 & 52.63 & \textbf{58.61} \\
& Individual & 68.09 & 58.18 & 16.67 & 7.41 & 42.86 & 72.73 & 42.54 & 46.11 & 44.25\\
 & Joint &  36.84 & 43.75 & 44.19 & 31.67 & 47.37 & 66.67 & 42.80 & 47.36 & 44.96\\
\midrule
\multirow{3}{*}{Qwen2.5-VL} & No & 74.36 & 26.36 & 28.57 & 90.67 & 86.67 & 12.15 & 63.20 & 43.06   & 55.09\\
& Individual& 63.64 & 50.91 & 22.50 & 33.33 & 64.52 & 60.61 & 50.22 & 48.28  & 49.23\\
 & Joint & 38.24 & 40.63 & 60.00 & 60.00 & 72.00 & 66.67 & 56.75 & 55.76   & \textbf{56.25}\\
\midrule
\multirow{3}{*}{LLaMA-3.2-Vision} & No & 62.50 & 13.51 & 29.33 & 91.67 & 66.67 & 8.00 & 52.83 & 37.73  & 44.02\\
& Individual & 76.19 & 29.09 & 33.33 & 22.22 & 38.16 & 87.88 & 49.23 & 46.40 & 47.77\\
 & Joint & 44.44 & 25.00 & 59.18 & 48.33 & 44.23 & 85.19 & 49.29 & 52.84  & \textbf{51.00}\\
\midrule
\multirow{3}{*}{Yi-VL} & No & 62.50 & 13.51 & 27.94 & 79.17 & 0.00 & 0.00 & 30.15 & 30.89 & 30.52 \\
& Individual& 36.84 & 12.73 & 35.29 & 44.44 & 29.03 & 54.55 & 33.72 & 37.24 & 35.39\\
 & Joint & 0.00 & 0.00 & 75.00 & 20.00 & 26.21 & 100.00 & 33.74 & 40.00 & \textbf{36.60 }\\
\midrule
\multirow{3}{*}{LLaVA-1.6} & No & 64.29 & 48.65 & 39.47 & 62.50 & 55.00 & 44.00 & 52.92 & 51.72 & 52.31\\
&Individual& 63.16 & 43.64 & 40.63 & 48.15 & 51.11 & 69.70 & 51.63 & 53.83 & 52.71\\
 & Joint & 58.06 & 56.25 & 70.00 & 46.67 & 45.83 & 81.48 & 57.97 & 61.47  & \textbf{59.66}\\
\bottomrule
\end{tabular}
}
\end{table*}

\begin{table*}[htbp]
\centering
\caption{Performance of Joint Fine-tuning with Type-II Visual Prompt for RailThreat Task on Cog-MRSI Dataset}
\label{tab:sota_mllm_type2-threat-MRSI-joint}
\renewcommand{\arraystretch}{1.3}
\resizebox{\textwidth}{!}{ \begin{tabular}{ccccccccccc}
\toprule
\multirow{2}{*}{\textbf{Model}} & \multirow{2}{*}{\textbf{Fine-tuned}} & \multicolumn{2}{c}{\textbf{Safe}} & \multicolumn{2}{c}{\textbf{Potential Threat}} & \multicolumn{2}{c}{\textbf{Serious Threat}} & \multicolumn{3}{c}{\textbf{Overall}} \\
\cline{3-11}
 & & Precision & Recall & Precision & Recall & Precision & Recall & Precision & Recall & \textbf{F1 Score} \\
\midrule
\multirow{3}{*}{Qwen2-VL} & No & 38.46 & 28.57 & 22.22 & 52.17 & 100.00 & 12.00 & 53.56 & 30.92 & 39.20 \\
& Individual & 68.75 & 60.00 & 52.17 & 44.44 & 52.27 & 69.70 & 57.73 & 58.05 & \textbf{57.89}\\
 & Joint &  72.00 & 42.86 & 20.00 & 26.67 & 51.61 & 84.21 & 47.87 & 51.24  & 49.50\\
\midrule
\multirow{3}{*}{Qwen2.5-VL} & No & 45.45 & 42.86 & 25.00 & 52.17 & 100.00 & 8.00 & 56.82 & 34.34   & 42.81\\
& Individual & 70.45 & 56.36 & 27.59 & 29.63 & 47.62 & 60.61 & 48.55 & 48.87  & 48.71\\
 & Joint & 74.19 & 54.76 & 22.22 & 26.67 & 51.85 & 73.68 & 49.42 & 51.70  & \textbf{50.54}\\
\midrule
\multirow{3}{*}{LLaMA-3.2-Vision} & No & 66.67 & 28.57 & 29.41 & 86.96 & 0.00 & 0.00 & 32.03 & 38.51  & 34.97\\
& Individual& 92.86 & 23.64 & 40.00 & 14.81 & 36.26 & 100.00 & 56.37 & 46.15 & 50.75\\
 & Joint & 80.00 & 47.62 & 31.58 & 40.00 & 56.25 & 94.74 & 55.94 & 60.79  & \textbf{58.26}\\
\midrule
\multirow{3}{*}{Yi-VL} & No & 25.00 & 8.57 & 29.82 & 73.91 & 0.00 & 0.00 & 18.27 & 27.49 & 21.96 \\
& Individual & 0.00 & 0.00 & 38.46 & 55.56 & 31.94 & 69.70 & 23.47 & 41.75 & 30.05\\
 & Joint & 57.14 & 28.57 & 0.00 & 0.00 & 29.09 & 84.21 & 28.74 & 37.59  & \textbf{32.58}\\
\midrule
\multirow{3}{*}{LLaVA-1.6} & No & 50.00 & 77.14 & 20.00 & 17.39 & 55.56 & 20.00 & 41.85 & 38.18 & 39.93\\
& Individual& 72.41 & 38.18 & 41.18 & 51.85 & 51.92 & 81.82 & 55.17 & 57.28 & \textbf{56.21}\\
 & Joint & 71.43 & 47.62 & 40.00 & 26.67 & 39.47 & 78.95 & 50.30 & 51.08   & 50.69\\
\bottomrule
\end{tabular}
}
\end{table*}

\begin{table*}[htbp]
\centering
\caption{Performance of Joint Fine-tuning with Type-I Visual Prompt for RailThreat Task on Cog-RailSem Dataset}
\label{tab:sota_mllm_type1-threat-RailSem-joint}
\renewcommand{\arraystretch}{1.3}
\resizebox{\textwidth}{!}{ \begin{tabular}{ccccccccccc}
\toprule
\multirow{2}{*}{\textbf{Model}} & \multirow{2}{*}{\textbf{Fine-tuned}} & \multicolumn{2}{c}{\textbf{Stationary}} & \multicolumn{2}{c}{\textbf{Threatening Move}} & \multicolumn{2}{c}{\textbf{Non-threatening Move}} & \multicolumn{3}{c}{\textbf{Overall}} \\
\cline{3-11}
 & & Precision & Recall & Precision & Recall & Precision & Recall & Precision & Recall & \textbf{F1 Score} \\
\midrule
\multirow{3}{*}{Qwen2-VL} & No & 50.00 & 63.16 & 33.33 & 28.57 & 62.07 & 52.94 & 48.47 & 48.22 & 48.35 \\
& Individual & 63.43 & 77.27 & 45.76 & 36.00 & 79.80 & 73.83 & 63.00 & 62.37 & 62.68\\
 & Joint &  69.44 & 80.65 & 76.67 & 63.89 & 79.41 & 81.82 & 75.17 & 75.45 & \textbf{75.31}\\
\midrule
\multirow{3}{*}{Qwen2.5-VL} & No & 51.72 & 39.47 & 31.67 & 54.29 & 66.67 & 35.29 & 50.02 & 43.02  & 46.26\\
& Individual & 65.45 & 65.45 & 45.83 & 44.00 & 70.00 & 71.96 & 60.43 & 60.47  & 60.45\\
 & Joint & 56.41 & 70.97 & 66.67 & 44.44 & 75.68 & 84.85 & 66.25 & 66.75 & \textbf{66.50}\\
\midrule
\multirow{3}{*}{LLaMA-3.2-Vision} & No & 28.00 & 18.42 & 32.43 & 68.57 & 83.33 & 14.71 & 47.92 & 33.90  & 39.71\\
& Individual & 63.64 & 76.36 & 56.10 & 30.67 & 71.43 & 79.44 & 63.72 & 62.16 & 62.93\\
 & Multi Task  & 54.76 & 74.19 & 73.68 & 38.89 & 69.23 & 81.82 & 65.89 & 64.97 & \textbf{65.43}\\
\midrule
\multirow{3}{*}{Yi-VL} & No & 27.27 & 7.89 & 36.99 & 77.14 & 66.67 & 5.88 & 43.64 & 30.31 & 35.77 \\
& Individual& 51.68 & 70.00 & 26.92 & 9.33 & 51.28 & 56.07 & 43.29 & 45.14 & 44.20\\
 & Joint & 41.30 & 61.29 & 47.62 & 27.78 & 51.52 & 51.52 & 46.81 & 46.86 & \textbf{46.84 }\\
\midrule
\multirow{3}{*}{LLaVA-1.6} & No & 40.32 & 65.79 & 28.57 & 22.86 & 52.94 & 26.47 & 40.61 & 38.37  & 39.46\\
& Individual& 60.56 & 78.18 & 54.84 & 22.67 & 57.14 & 63.55 & 57.51 & 54.80 & 56.12\\
 & Joint & 60.47 & 83.87 & 71.43 & 41.67 & 66.67 & 72.73 & 66.19 & 66.09 & \textbf{66.14}\\
\bottomrule
\end{tabular}
}
\end{table*}

\begin{table*}[htbp]
\centering
\caption{Performance of Joint Fine-tuning with Type-II Visual Prompt for RailThreat Task on Cog-RailSem Dataset}
\label{tab:sota_mllm_type2-threat-RailSem-joint}
\renewcommand{\arraystretch}{1.3}
\resizebox{\textwidth}{!}{ \begin{tabular}{ccccccccccc}
\toprule
\multirow{2}{*}{\textbf{Model}} & \multirow{2}{*}{\textbf{Fine-tuned}} & \multicolumn{2}{c}{\textbf{Stationary}} & \multicolumn{2}{c}{\textbf{Threatening Move}} & \multicolumn{2}{c}{\textbf{Non-threatening Move}} & \multicolumn{3}{c}{\textbf{Overall}} \\
\cline{3-11}
 & & Precision & Recall & Precision & Recall & Precision & Recall & Precision & Recall & \textbf{F1 Score} \\
\midrule
\multirow{3}{*}{Qwen2-VL} & No & 47.83 & 62.86 & 33.33 & 31.71 & 50.00 & 37.84 & 43.72 & 44.13 & 43.93 \\
& Individual & 65.63 & 76.36 & 58.00 & 38.67 & 75.44 & 80.37 & 66.35 & 65.13 & 65.74\\
 & Joint &  83.78 & 81.58 & 60.87 & 60.87 & 75.00 & 77.42 & 73.22 & 73.29 & \textbf{73.25}\\
\midrule
\multirow{3}{*}{Qwen2.5-VL} & No & 54.05 & 57.14 & 43.94 & 70.73 & 90.00 & 24.32 & 62.66 & 50.73   & 56.07\\
& Individual & 66.38 & 70.00 & 54.93 & 52.00 & 78.10 & 76.64 & 66.47 & 66.21  & 66.34\\
 & Joint & 85.71 & 78.95 & 59.09 & 56.52 & 82.86 & 93.55 & 75.89 & 76.34 & \textbf{76.11}\\
\midrule
\multirow{3}{*}{LLaMA-3.2-Vision} & No & 33.33 & 25.71 & 33.33 & 65.85 & 0.00 & 0.00 & 22.22 & 30.52  & 25.72\\
& Individual & 67.42 & 80.91 & 68.09 & 42.67 & 74.34 & 78.50 & 69.95 & 67.36 & 68.63\\
 & Joint & 81.58 & 81.58 & 68.75 & 47.83 & 76.32 & 93.55 & 75.55 & 74.32 & \textbf{74.93}\\
\midrule
\multirow{3}{*}{Yi-VL} & No & 50.00 & 8.57 & 37.65 & 78.05 & 100.00 & 2.70 & 62.55 & 29.77   & 40.34 \\
& Individual& 49.70 & 74.55 & 42.86 & 4.00 & 50.00 & 56.07 & 47.52 & 44.87  & 46.16\\
 & Joint & 64.86 & 63.16 & 40.00 & 17.39 & 51.11 & 74.19 & 51.99 & 51.58 & \textbf{51.79}\\
\midrule
\multirow{3}{*}{LLaVA-1.6} & No & 39.62 & 60.00 & 34.38 & 26.83 & 25.93 & 18.92 & 33.31 & 35.25 & 34.25\\
& Individual& 64.71 & 70.00 & 77.27 & 22.67 & 56.29 & 79.44 & 66.09 & 57.37  & 61.42\\
 & Joint & 78.38 & 76.32 & 60.00 & 26.09 & 60.00 & 87.10 & 66.13 & 63.17 & \textbf{64.61}\\
\bottomrule
\end{tabular}
}
\end{table*}

In this section, we conduct systematic evaluations on RailGPT based on five state-of-the-art (SOTA) multimodal large models: Qwen2-VL-7B-Instruct (denoted as Qwen2-VL) \cite{wang2024qwen2vlenhancingvisionlanguagemodels}, Qwen2.5-VL-7B-Instruct (denoted as Qwen2.5-VL) \cite{bai2025qwen25vltechnicalreport}, LLaMA-3.2-11B-Vision-Instruct (denoted as LLaMA-3.2-Vision) \cite{grattafioriLlamaHerdModels2024}, Yi-VL-6B (denoted as Yi-VL) \cite{ai2025yiopenfoundationmodels}, and LLaVA-NeXT-7B (denoted as LLaVA-1.6) \cite{liu2024llavanext}. We adopt a 4:1 training–test split for both Cog‑MRSI and Cog‑RailSem19. All tasks are formulated as single‑answer multiple‑choice, i.e., each question provides several options, and the model must output “Answer: X” (with X denoting the selected option), enabling automatic scoring. To mitigate bias from uneven answer distributions in the test set, we compute precision and recall for each of the three answer types and their averages, and report the harmonic mean (F1) as the aggregate metric. For occasional format violations, we apply a semantic‑similarity fallback: the option with the highest weighted combination of BLEU, ROUGE‑1/‑2, and cosine similarity to the model output is selected as the prediction, preserving completeness and comparability.

\subsection{Performance of SOTA VLMs on CogRail}

\begin{figure}[htbp]
\centering
\includegraphics[width=3in]{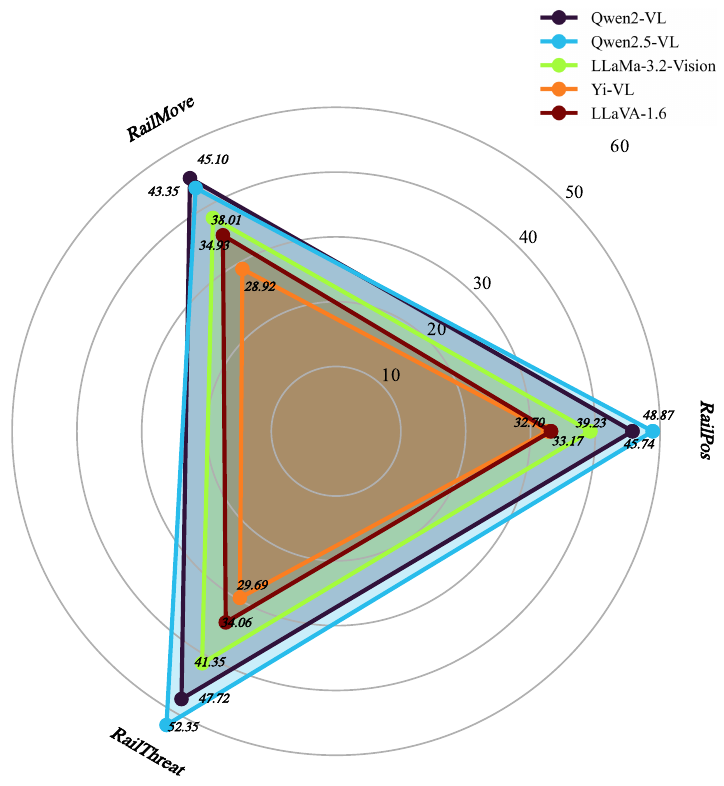}
\caption{Performance Comparison among SOTA VLMs on CogRail averaged on different prompt types and sub-datasets}
\label{sota_average_wo_ft}
\end{figure}

In the zero‑shot setting, i.e., without any task‑specific fine‑tuning and relying solely on publicly released checkpoints, we evaluate five leading vision–language models: Qwen2‑VL, Qwen2.5‑VL, LLaMA‑3.2‑Vision, Yi‑VL, and LLaVA‑1.6. Each model was tested on the three CogRail subtasks (RailPOS, RailMove, and RailThreat) using both Type‑I and Type‑II prompt templates. Performance scores reported in Figure \ref{sota_average_wo_ft} represent the average accuracy across prompt types and the Cog‑MRSI and Cog‑RailSem19 datasets. Among the five, Qwen2.5‑VL attains the highest overall accuracy (48.19\%), achieving 48.87\% on RailPOS and 52.35\% on RailThreat. Qwen2‑VL led on RailMove with 45.10\%, outperforming Qwen2.5‑VL’s 43.35\%. LLaMA‑3.2‑Vision, LLaVA‑1.6, and Yi‑VL reach mean accuracies of 39.53\%, 34.05\%, and 30.44\%, respectively. Notably, Qwen2.5‑VL outperforms Yi‑VL by 22.66 percentage points on RailThreat. Although absolute accuracies differ by task, model rankings remain consistent, indicating that architectural enhancements and expanded pre‑training corpora uniformly benefit localization, motion reasoning, and risk assessment\cite{bai2025qwen25vltechnicalreport}.

\subsection{Individual Fine-tuning of SOTA VLMs}

To further enhance model performance across the three subtasks defined in CogRail, we conducted individual, task-specific fine-tuning of each SOTA open-source vision-language model (VLM) using multimodal instruction-based supervision. For each subtask, the models were fine-tuned separately on the \textit{Cog-MRSI} and \textit{Cog-RailSem19} datasets under both Type-I and Type-II prompt formulations. After fine-tuning, most models exhibited substantial improvements in F1 score, as summarized in Figures~\ref{ex_data_show_1}–\ref{ex_data_show_4}. The final subplot in each figure reports the average performance across the five evaluated models. Detailed numerical results corresponding to each configuration are provided in the Appendix for reference.

In the \textit{RailPOS} task, for example, under the Type-I prompt setting on the Cog-MRSI dataset, the fine-tuned Qwen2.5-VL achieved the highest F1 score of 51.41\%, while LLaMA-3.2-Vision achieved the most significant improvement, with F1 increasing from 35.99\% to 50.18\%. Averaged across all five models, fine-tuning led to substantial F1-score improvements under the four experimental settings (Figures~\ref{ex_data_show_1}–\ref{ex_data_show_4}), with increases of 7.47\% (from 29.90\% to 37.37\%), 4.77\% (from 35.51\% to 40.28\%), 18.35\% (from 43.00\% to 61.35\%), and 10.36\% (from 44.65\% to 55.01\%), respectively.

In the \textit{RailMove} task, under the Type-II prompt on the Cog-MRSI dataset, LLaMA-3.2-Vision achieved the best overall F1 score at 55.38\%, while LLaVa-1.6 demonstrated the greatest improvement, with F1 increasing from 32.57\% to 55.15\%, a gain of 22.58 points. On average, models improved by 10.10 points (from 36.93\% to 47.03\%).

In the \textit{RailThreat} task, using the Type-II visual prompt on the Cog-RailSem19 dataset, LLaMA-3.2-Vision again achieved the highest post-fine-tuning F1 score of 68.63\%, also representing the largest performance gain among all models in this task, with a 17.41-point improvement from 51.22\%. The average improvement across the five models was 18.60 points (from 43.06\% to 61.66\%).

Although overall performance was significantly enhanced after fine-tuning, we also observed several counterintuitive cases. For instance, Qwen2.5-VL and Yi-VL exhibited performance degradation under certain RailMove and RailThreat configurations. This may be partially attributed to the uniform fine-tuning strategy applied across all models—including identical data splits and hyperparameters—which might not be optimal for each model architecture. More importantly, these cases reveal the inherent complexity of cognitive railway perception tasks. In particular, tasks such as motion inference and threat estimation based solely on single-frame visual input are intrinsically ambiguous and cognitively demanding, highlighting the need for richer context modeling and temporal understanding in future research.

\subsection{Joint Fine-tuning of SOTA VLMs}

To further investigate the benefits of multi-task supervision for VLM fine-tuning, we designed a joint fine-tuning scheme in which the models were trained simultaneously from all three task categories. The corresponding results are presented in Tables~\ref{tab:sota_mllm_type1-threat-MRSI-joint}–\ref{tab:sota_mllm_type2-threat-RailSem-joint}. Among the 20 experimental configurations (comprising different models, prompt types, and datasets), joint fine-tuning outperformed both the zero-shot and individually fine-tuned baselines in 17 cases, demonstrating its effectiveness and practical value. These improvements are largely due to the richer supervision signals provided by joint training, as well as the interrelated nature of the three sub-tasks, which supports more effective knowledge transfer and enhances the model's ability to recognize intrusion threats.

Specifically, for the Type-I visual prompt on the RailThreat task using the Cog-MRSI dataset, LLaVA-1.6 achieved the best performance with an F1 score of 59.66\%, representing a 6.95-point gain over individual fine-tuning. On the Type-II visual prompt for the same task and dataset, LLaMA-3.2-Vision reached 58.26\%, yielding improvements of 23.29 points over the zero-shot baseline and 7.51 points over individual fine-tuning. For the Type-I prompt on the Cog-RailSem19 dataset, Qwen2-VL achieved the best result with an F1 score of 75.31\%, surpassing the no-tuning and individual tuning settings by 26.96 and 12.63 points, respectively. Finally, under the Type-II prompt on the same dataset, Qwen2.5-VL achieved the highest F1 score of 76.11\%, reflecting gains of 20.04 and 9.77 points over the zero-shot and individually tuned versions.

\section{Conclusion}
\label{sec:conclusion}
This paper presents CogRail, the first benchmark for railway intrusion perception that integrates spatial positioning, motion state, and threat assessment into a unified task system. It provides a standardized foundation for training and evaluating multimodal models in safety-critical railway scenarios. Building upon CogRail, we propose RailGPT, a flexible, agent-based multimodal framework designed to support diverse vision-language models through tailored prompting and fine-tuning strategies. By incorporating both individual and joint fine-tuning across tasks, RailGPT achieves accurate and interpretable reasoning for intrusion detection. Experiments demonstrate significant improvements across all subtasks, \textit{RailPos}, \textit{RailMove}, and \textit{RailThreat}, highlighting the value of multi-task learning and the potential of foundation models in domain-specific cognitive reasoning. Future work will explore long-horizon reasoning and robust adaptation in challenging operational environments.





\ifCLASSOPTIONcaptionsoff
  \newpage
\fi



%
\bibliographystyle{IEEEtran}
\bibliography{railbib}

\newpage
\newpage

\end{document}


%
\title{Supplementary Material}
%
%
%

\author{Yonglin~Tian$^{1}^{*}$, ~\IEEEmembership{Member,~IEEE}, Qiyao~Zhang$^{2}^{*}$, Wei~Xu$^{3}$, Yutong~Wang$^{1}$, Yihao~Wu$^{4}$,  Xinyi~Li$^{4}$, Hui~Zhang$^{5}$, Zhiyong~Cui$^{1}$, Baoqing~Guo$^{6}$, ~\IEEEmembership{Member,~IEEE}, Yisheng~Lv$^{1}$, ~\IEEEmembership{Member,~IEEE,} \\Zujun~Yu$^{6}$, and Fei-Yue Wang$^{7}$,~\IEEEmembership{Fellow,~IEEE}
}

\markboth{Journal of \LaTeX\ Class Files,~Vol.~14, No.~8, August~2015}%
{Shell \MakeLowercase{\textit{et al.}}: Bare Demo of IEEEtran.cls for IEEE Journals}

\maketitle

\IEEEpeerreviewmaketitle

In the supplementary material, we provide comprehensive tabulations of the individual fine‑tuning results for all five VLMs across both the Cog‑MRSI and Cog‑RailSem19 datasets under Type-I and Type-II visual prompts. Tables \ref{tab:sota_mllm_type1_pos_MRSI}–\ref{tab:sota_mllm_type1_threat_MRSI} report per‑class precision, recall, and overall F1 for the RailPos, RailMove, and RailThreat tasks on Cog‑MRSI with Type-I prompts; Tables \ref{tab:sota_mllm_type2_pos_MRSI}–\ref{tab:sota_mllm_type2_threat_MRSI} report the results for Type-II prompts. Analogously, Tables \ref{tab:sota_mllm_type1_pos_RailSem19}–\ref{tab:sota_mllm_type1_threat_RailSem} cover the Cog‑RailSem19 dataset with Type-I prompts, while Tables \ref{tab:sota_mllm_type2_pos_RailSem19}–\ref{tab:sota_mllm_type2_threat_RailSem} present the Type-II results. These appended tables enable detailed comparison of task‑ and class‑level gains, complementing the averaged performance curves shown in Figures 5–8 of the main text.

\begin{table*}[htbp]
\centering
\caption{Performance Comparison via Individual Fine-Tuning with Type-I Visual Prompt on RailPos Task in Cog-MRSI Dataset}
\label{tab:sota_mllm_type1_pos_MRSI}
\renewcommand{\arraystretch}{1.3}
\resizebox{\textwidth}{!}{ \begin{tabular}{ccccccccccc}
\toprule
\multirow{2}{*}{\textbf{Model}} & \multirow{2}{*}{\textbf{Fine-tuned}} & \multicolumn{2}{c}{\textbf{On Track}} & \multicolumn{2}{c}{\textbf{On Ballast}} & \multicolumn{2}{c}{\textbf{Outside Ballast}} & \multicolumn{3}{c}{\textbf{Overall}} \\
\cline{3-11}
 & & Precision & Recall & Precision & Recall & Precision & Recall & Precision & Recall & \textbf{F1 Score} \\
\hline
\multirow{2}{*}{Qwen2-VL} & No & 88.24 & 18.52 & 3.57 & 14.29 & 34.29 & 88.89 & 42.03 & 40.56 & 41.28\\
 & Yes & 86.21 & 61.73 & 6.67 & 28.57 & 55.56 & 55.56 & 49.48 & 48.62 & \textbf{49.05} \\
\hline
\multirow{2}{*}{Qwen2.5-VL} & No & 78.05 & 39.51 & 6.56 & 57.14 & 76.92 & 37.04 & 53.84 & 44.56 & 48.76\\
 & Yes & 85.25 & 64.20 & 10.71 & 42.86 & 53.85 & 51.85 & 49.94 & 52.97 & \textbf{51.41}\\
\hline
\multirow{2}{*}{Llama-3.2-Vision} & No & 78.95 & 18.52 & 0.00 & 0.00 & 26.04 & 92.59 & 35.00 & 37.04 & 35.99\\
 & Yes & 87.50 & 17.28 & 14.29 & 57.14 & 35.21 & 92.59 & 45.67 & 55.67 & \textbf{50.18}\\
\hline
\multirow{2}{*}{Yi-VL} & No & 70.27 & 96.30 & 0.00 & 0.00 & 0.00 & 0.00 & 23.42 & 32.10 & 27.08\\
 & Yes & 33.33 & 2.47 & 9.68 & 85.71 & 23.40 & 40.74 & 22.14 & 42.97 & \textbf{29.22}\\
\hline
\multirow{2}{*}{LLaVA-1.6} & No & 73.02 & 56.79 & 2.56 & 14.29 & 23.08 & 11.11 & 32.89 & 27.40 & 29.90\\
 & Yes & 76.19 & 79.01 & 0.00 & 0.00 & 32.26 & 37.04 & 36.15 & 38.68 & \textbf{37.37}\\
\bottomrule
\end{tabular}}
\end{table*}

\begin{table*}[htbp]
\centering
\caption{Performance Comparison via Individual Fine-Tuning with Type-I Visual Prompt on RailMove Task in Cog-MRSI Dataset}
\label{tab:sota_mllm_type1_move_MRSI}
\renewcommand{\arraystretch}{1.3}
\resizebox{\textwidth}{!}{ \begin{tabular}{ccccccccccc}
\toprule
\multirow{2}{*}{\textbf{Model}} & \multirow{2}{*}{\textbf{Fine-tuned}} & \multicolumn{2}{c}{\textbf{Stationary}} & \multicolumn{2}{c}{\textbf{Threatening Move}} & \multicolumn{2}{c}{\textbf{Non-threatening Move}} & \multicolumn{3}{c}{\textbf{Overall}} \\
\cline{3-11}
 & & Precision & Recall & Precision & Recall & Precision & Recall & Precision & Recall & \textbf{F1 Score} \\
\midrule
\multirow{2}{*}{Qwen2-VL} & No & 33.33 & 66.67 & 35.71 & 12.82 & 64.62 & 72.41 & 44.55 & 50.63 &47.40 \\
 & Yes & 40.62 & 72.22 & 52.63 & 25.64 & 67.19 & 74.14 & 53.48 & 57.33 & \textbf{55.34}\\
\midrule
\multirow{2}{*}{Qwen2.5-VL} & No & 25.00 & 50.00 & 37.31 & 64.10 & 66.67 & 13.79 & 42.99 & 42.63 & 42.81\\
 & Yes & 25.00 & 33.33 & 47.37 & 46.15 & 64.15 & 58.62 & 45.51 & 46.04 & \textbf{45.77}\\
\midrule
\multirow{2}{*}{Llama-3.2-Vision} & No & 18.75 & 16.67 & 32.18 & 71.79 & 58.33 & 12.07 & 36.42 & 33.51 & 34.90\\
 & Yes & 36.00 & 50.00 & 50.00 & 35.90 & 70.97 & 75.86 & 52.32 & 53.92 & \textbf{53.11}\\
\midrule
\multirow{2}{*}{Yi-VL} & No & 12.35 & 55.56 & 35.29 & 30.77 & 0.00 & 0.00 & 15.88 & 28.77 & 20.46\\
 & Yes & 0.00 & 0.00 & 31.03 & 32.08 & 50.00 & 63.79 & 27.01 & 28.96 & \textbf{27.95}\\
\midrule
\multirow{2}{*}{LLaVA-1.6} & No & 31.25 & 55.56 & 37.50 & 38.46 & 56.25 & 31.03 & 41.67 & 41.68 & 41.67\\
 & Yes & 50.00 & 55.56 & 36.36 & 10.26 & 55.95 & 81.03 & 47.44 & 48.95 & \textbf{48.18}\\
\bottomrule
\end{tabular}}
\end{table*}

\begin{table*}[htbp]
\centering
\caption{Performance Comparison via Individual Fine-Tuning with Type-I Visual Prompt on RailThreat Task in Cog-MRSI Dataset}
\label{tab:sota_mllm_type1_threat_MRSI}
\renewcommand{\arraystretch}{1.3}
\resizebox{\textwidth}{!}{ \begin{tabular}{ccccccccccc}
\toprule
\multirow{2}{*}{\textbf{Model}} & \multirow{2}{*}{\textbf{Fine-tuned}} & \multicolumn{2}{c}{\textbf{Safe}} & \multicolumn{2}{c}{\textbf{Potential Threat}} & \multicolumn{2}{c}{\textbf{Serious Threat}} & \multicolumn{3}{c}{\textbf{Overall}} \\
\cline{3-11}
 & & Precision & Recall & Precision & Recall & Precision & Recall & Precision & Recall & \textbf{F1 Score} \\
\midrule
\multirow{2}{*}{Qwen2-VL} & No & 60.00 & 54.55 & 24.14 & 51.85 & 85.71 & 18.18 & 56.62 & 41.53 & \textbf{47.91} \\
 & Yes &  68.09 & 58.18 & 16.67 & 7.41 & 42.86 & 72.73 & 42.54 & 46.11 & 44.25\\
\midrule
\multirow{2}{*}{Qwen2.5-VL} & No & 84.21 & 29.09 & 28.26 & 96.30 & 100.00 & 12.12 & 70.82 & 45.84  & \textbf{55.65}\\
 & Yes & 63.64 & 50.91 & 22.50 & 33.33 & 64.52 & 60.61 & 50.22 & 48.28  & 49.23\\
\midrule
\multirow{2}{*}{Llama-3.2-Vision} & No & 100.00 & 7.27 & 24.51 & 92.59 & 50.00 & 12.12 & 58.17 & 37.33  & 45.48\\
 & Yes & 76.19 & 29.09 & 33.33 & 22.22 & 38.16 & 87.88 & 49.23 & 46.40 & \textbf{47.77}\\
\midrule
\multirow{2}{*}{Yi-VL} & No & 41.67 & 9.09 & 23.16 & 81.48 & 0.00 & 0.00 & 21.61 & 30.19 & 25.19 \\
 & Yes & 36.84 & 12.73 & 35.29 & 44.44 & 29.03 & 54.55 & 33.72 & 37.24 & \textbf{35.39}\\
\midrule
\multirow{2}{*}{LLaVA-1.6} & No & 62.50 & 18.18 & 24.72 & 81.48 & 20.00 & 6.06 & 35.74 & 35.24  & 35.49\\
 & Yes &  63.16 & 43.64 & 40.63 & 48.15 & 51.11 & 69.70 & 51.63 & 53.83 & \textbf{52.71}\\
\bottomrule
\end{tabular}}
\end{table*}

\begin{table*}[htbp]
\centering
\caption{Performance Comparison via Individual Fine-Tuning with Type-II Visual Prompt on RailPos Task in Cog-MRSI Dataset}
\label{tab:sota_mllm_type2_pos_MRSI}
\renewcommand{\arraystretch}{1.3}
\resizebox{\textwidth}{!}{ \begin{tabular}{ccccccccccc}
\toprule
\multirow{2}{*}{\textbf{Model}} & \multirow{2}{*}{\textbf{Fine-tuned}} & \multicolumn{2}{c}{\textbf{On Track}} & \multicolumn{2}{c}{\textbf{On Ballast}} & \multicolumn{2}{c}{\textbf{Outside Ballast}} & \multicolumn{3}{c}{\textbf{Overall}} \\
\cline{3-11}
 & & Precision & Recall & Precision & Recall & Precision & Recall & Precision & Recall & \textbf{F1 Score} \\
\hline
\multirow{2}{*}{Qwen2-VL} & No & 100.00 & 9.88 & 2.44 & 14.29 & 34.85 & 85.19 & 45.76 & 36.45  &40.58\\
 & Yes & 80.33 & 60.49 & 5.56 & 28.57 & 50.00 & 33.33 & 45.29 & 40.80  & \textbf{42.93} \\
\hline
\multirow{2}{*}{Qwen2.5-VL} & No & 85.71 & 29.63 & 6.41 & 71.43 & 55.56 & 18.52 & 49.23 & 39.86 & \textbf{44.05}\\
 & Yes & 85.45 & 58.02 & 3.57 & 14.29 & 43.75 & 51.85 & 44.26 & 41.39 & 42.78\\
\hline
\multirow{2}{*}{Llama-3.2-Vision} & No & 75.00 & 29.63 & 0.00 & 0.00 & 28.75 & 85.19 & 34.58 & 38.27  & 36.33\\
 & Yes & 87.18 & 41.98 & 12.00 & 42.86 & 43.14 & 81.48 & 47.44 & 55.44  & \textbf{51.13}\\
\hline
\multirow{2}{*}{Yi-VL} & No & 69.72 & 93.83 & 0.00 & 0.00 & 0.00 & 0.00 & 23.24 & 31.28 & \textbf{26.67}\\
 & Yes & 42.11 & 9.88 & 0.00 & 0.00 & 22.22 & 51.85 & 21.44 & 20.58  & 21.00\\
\hline
\multirow{2}{*}{LLaVA-1.6} & No & 73.02 & 56.79 & 2.56 & 14.29 & 23.08 & 11.11 & 32.89 & 27.40 & 29.90\\
 & Yes & 80.00 & 69.14 & 16.67 & 28.57 & 30.30 & 37.04 & 42.32 & 44.91 & \textbf{43.58}\\
\bottomrule
\end{tabular}}
\end{table*}

\begin{table*}[htbp]
\centering
\caption{Performance Comparison via Individual Fine-Tuning with Type-II Visual Prompt on RailMove Task in Cog-MRSI Dataset}
\label{tab:sota_mllm_type2_move_MRSI}
\renewcommand{\arraystretch}{1.3}
\resizebox{\textwidth}{!}{ \begin{tabular}{ccccccccccc}
\toprule
\multirow{2}{*}{\textbf{Model}} & \multirow{2}{*}{\textbf{Fine-tuned}} & \multicolumn{2}{c}{\textbf{Stationary}} & \multicolumn{2}{c}{\textbf{Threatening Move}} & \multicolumn{2}{c}{\textbf{Non-threatening Move}} & \multicolumn{3}{c}{\textbf{Overall}} \\
\cline{3-11}
 & & Precision & Recall & Precision & Recall & Precision & Recall & Precision & Recall & \textbf{F1 Score} \\
\midrule
\multirow{2}{*}{Qwen2-VL} & No & 28.57 & 44.44 & 40.00 & 15.28 & 61.11 & 75.86 & 43.23 & 45.23 &44.21 \\
 & Yes &  37.93 & 61.11 & 50.00 & 25.64 & 63.64 & 72.41 & 50.52 & 53.06 & \textbf{51.76}\\
\midrule
\multirow{2}{*}{Qwen2.5-VL} & No & 7.69 & 5.56 & 33.33 & 79.49 & 88.89 & 13.79 & 43.30 & 32.95 & 37.42\\
 & Yes & 15.38 & 11.11 & 39.29 & 56.41 & 67.39 & 53.45 & 40.69 & 50.32 & \textbf{45.00}\\
\midrule
\multirow{2}{*}{Llama-3.2-Vision} & No & 36.36 & 22.22 & 35.37 & 74.36 & 68.18 & 25.86 & 46.64 & 40.81 & 43.53\\
 & Yes & 50.00 & 22.22 & 52.50 & 53.85 & 71.64 & 82.76 & 58.05 & 52.94 & \textbf{55.38}\\
\midrule
\multirow{2}{*}{Yi-VL} & No & 9.78 & 50.00 & 25.00 & 10.26 & 42.86 & 5.17 & 25.88 & 28.06 & 26.92 \\
 & Yes & 0.00 & 0.00 & 31.11 & 35.90 & 51.85 & 48.28 & 27.65 & 28.06 & \textbf{27.85}\\
\midrule
\multirow{2}{*}{LLaVA-1.6} & No & 19.44 & 38.89 & 31.82 & 35.90 & 47.83 & 18.97 & 33.03 & 32.12 & 32.57\\
 & Yes & 42.11 & 44.44 & 57.69 & 38.46 & 67.14 & 81.03 & 55.65 & 54.65 & \textbf{55.15}\\
\bottomrule
\end{tabular}}
\end{table*}

\begin{table*}[htbp]
\centering
\caption{Performance Comparison via Individual Fine-Tuning with Type-II Visual Prompt on RailThreat Task in Cog-MRSI Dataset}
\label{tab:sota_mllm_type2_threat_MRSI}
\renewcommand{\arraystretch}{1.3}
\resizebox{\textwidth}{!}{ \begin{tabular}{ccccccccccc}
\toprule
\multirow{2}{*}{\textbf{Model}} & \multirow{2}{*}{\textbf{Fine-tuned}} & \multicolumn{2}{c}{\textbf{Safe}} & \multicolumn{2}{c}{\textbf{Potential Threat}} & \multicolumn{2}{c}{\textbf{Serious Threat}} & \multicolumn{3}{c}{\textbf{Overall}} \\
\cline{3-11}
 & & Precision & Recall & Precision & Recall & Precision & Recall & Precision & Recall & \textbf{F1 Score} \\
\midrule
\multirow{2}{*}{Qwen2-VL} & No & 78.95 & 27.27 & 25.81 & 88.89 & 100.00 & 9.09 & 68.25 & 41.75   & 51.81 \\
 & Yes &  68.75 & 60.00 & 52.17 & 44.44 & 52.27 & 69.70 & 57.73 & 58.05 & \textbf{57.89}\\
\midrule
\multirow{2}{*}{Qwen2.5-VL} & No & 70.00 & 25.45 & 24.44 & 81.48 & 100.00 & 15.15 & 64.81 & 40.70   & \textbf{50.00}\\
 & Yes & 70.45 & 56.36 & 27.59 & 29.63 & 47.62 & 60.61 & 48.55 & 48.87  & 48.71\\
\midrule
\multirow{2}{*}{Llama-3.2-Vision} & No & 70.00 & 12.73 & 23.08 & 77.78 & 42.86 & 18.18 & 45.31 & 36.23  & 40.26\\
 & Yes & 92.86 & 23.64 & 40.00 & 14.81 & 36.26 & 100.00 & 56.37 & 46.15 & \textbf{50.75}\\
\midrule
\multirow{2}{*}{Yi-VL} & No & 22.22 & 7.27 & 18.39 & 59.26 & 0.00 & 0.00 & 13.54 & 22.18  & 16.81 \\
 & Yes & 0.00 & 0.00 & 38.46 & 55.56 & 31.94 & 69.70 & 23.47 & 41.75 & \textbf{30.05}\\
\midrule
\multirow{2}{*}{LLaVA-1.6} & No & 50.00 & 21.82 & 23.81 & 74.07 & 14.29 & 3.03 & 29.37 & 32.97  & 31.07\\
 & Yes &  72.41 & 38.18 & 41.18 & 51.85 & 51.92 & 81.82 & 55.17 & 57.28 & \textbf{56.21}\\
\bottomrule
\end{tabular}}
\end{table*}

\begin{table*}[htbp]
\centering
\caption{Performance Comparison via Individual Fine-Tuning with Type-I Visual Prompt on RailPos Task in Cog-RailSem19 Dataset}
\label{tab:sota_mllm_type1_pos_RailSem19}
\renewcommand{\arraystretch}{1.3}
\resizebox{\textwidth}{!}{ \begin{tabular}{ccccccccccc}
\toprule
\multirow{2}{*}{\textbf{Model}} & \multirow{2}{*}{\textbf{Fine-tuned}} & \multicolumn{2}{c}{\textbf{On Track}} & \multicolumn{2}{c}{\textbf{On Ballast}} & \multicolumn{2}{c}{\textbf{Outside Ballast}} & \multicolumn{3}{c}{\textbf{Overall}} \\
\cline{3-11}
 & & Precision & Recall & Precision & Recall & Precision & Recall & Precision & Recall & \textbf{F1 Score} \\
\hline
\multirow{2}{*}{Qwen2-VL} & No & 92.86 & 16.25 & 13.83 & 30.23 & 44.71 & 85.39 & 50.46 & 43.96 & 46.99\\
 & Yes & 85.16 & 82.50 & 35.48 & 51.16 & 80.00 & 67.42 & 66.88 & 67.03 & \textbf{66.95} \\
\hline
\multirow{2}{*}{Qwen2.5-VL} & No & 83.33 & 37.50 & 13.82 & 39.53 & 61.86 & 67.42 & 53.00 & 48.15 & 50.46\\
 & Yes & 85.06 & 81.88 & 26.67 & 37.21 & 70.51 & 61.80 & 60.75 & 60.29 & \textbf{60.52}\\
\hline
\multirow{2}{*}{Llama-3.2-Vision} & No & 60.87 & 70.00 & 25.00 & 9.30 & 43.33 & 43.82 & 43.07 & 41.04  & 42.03\\
 & Yes & 85.12 & 89.38 & 55.81 & 55.81 & 80.25 & 73.03 & 73.73 & 72.74 & \textbf{73.23}\\
\hline
\multirow{2}{*}{Yi-VL} & No & 55.43 & 89.38 & 23.33 & 16.28 & 50.00 & 2.25 & 42.92 & 35.97  & 39.14\\
 & Yes & 66.67 & 65.00 & 24.53 & 30.23 & 53.01 & 49.44 & 48.07 & 48.22 & \textbf{48.15}\\
\hline
\multirow{2}{*}{LLaVA-1.6} & No &57.55 & 50.00 & 17.60 & 51.16 & 35.29 & 6.74 & 36.82 & 35.97 & 36.39\\
 & Yes & 78.26 & 67.50 & 32.88 & 55.81 & 59.26 & 53.93 & 56.80 & 59.08 & \textbf{57.92}\\
\bottomrule
\end{tabular}}
\end{table*}

\begin{table*}[htbp]
\centering
\caption{Performance Comparison via Individual Fine-Tuning with Type-I Visual Prompt on RailMove Task in Cog-RailSem19 Dataset}
\label{tab:sota_mllm_type1_move_RailSem19}
\renewcommand{\arraystretch}{1.3}
\resizebox{\textwidth}{!}{ \begin{tabular}{ccccccccccc}
\toprule
\multirow{2}{*}{\textbf{Model}} & \multirow{2}{*}{\textbf{Fine-tuned}} & \multicolumn{2}{c}{\textbf{Stationary}} & \multicolumn{2}{c}{\textbf{Threatening Move}} & \multicolumn{2}{c}{\textbf{Non-threatening Move}} & \multicolumn{3}{c}{\textbf{Overall}} \\
\cline{3-11}
 & & Precision & Recall & Precision & Recall & Precision & Recall & Precision & Recall & \textbf{F1 Score} \\
\hline
\multirow{2}{*}{Qwen2-VL} & No & 100.00 & 9.88 & 2.44 & 14.29 & 34.85 & 85.19 & 45.76 & 36.45  &40.58\\
 & Yes & 80.33 & 60.49 & 5.56 & 28.57 & 50.00 & 33.33 & 45.29 & 40.80  & \textbf{42.93} \\
\hline
\multirow{2}{*}{Qwen2.5-VL} & No & 85.71 & 29.63 & 6.41 & 71.43 & 55.56 & 18.52 & 49.23 & 39.86 & \textbf{44.05}\\
 & Yes & 85.45 & 58.02 & 3.57 & 14.29 & 43.75 & 51.85 & 44.26 & 41.39 & 42.78\\
\hline
\multirow{2}{*}{Llama-3.2-Vision} & No & 75.00 & 29.63 & 0.00 & 0.00 & 28.75 & 85.19 & 34.58 & 38.27  & 36.33\\
 & Yes & 87.18 & 41.98 & 12.00 & 42.86 & 43.14 & 81.48 & 47.44 & 55.44  & \textbf{51.13}\\
\hline
\multirow{2}{*}{Yi-VL} & No & 69.72 & 93.83 & 0.00 & 0.00 & 0.00 & 0.00 & 23.24 & 31.28 & \textbf{26.67}\\
 & Yes & 42.11 & 9.88 & 0.00 & 0.00 & 22.22 & 51.85 & 21.44 & 20.58  & 21.00\\
\hline
\multirow{2}{*}{LLaVA-1.6} & No & 73.02 & 56.79 & 2.56 & 14.29 & 23.08 & 11.11 & 32.89 & 27.40 & 32.05\\
 & Yes & 80.00 & 69.14 & 16.67 & 28.57 & 30.30 & 37.04 & 42.32 & 44.91 & \textbf{43.58}\\
\bottomrule
\end{tabular}}
\end{table*}

\begin{table*}[htbp]
\centering
\caption{Performance Comparison via Individual Fine-Tuning with Type-I Visual Prompt on RailThreat Task in Cog-RailSem Dataset}
\label{tab:sota_mllm_type1_threat_RailSem}
\renewcommand{\arraystretch}{1.3}
\resizebox{\textwidth}{!}{ \begin{tabular}{ccccccccccc}
\toprule
\multirow{2}{*}{\textbf{Model}} & \multirow{2}{*}{\textbf{Fine-tuned}} & \multicolumn{2}{c}{\textbf{Safe}} & \multicolumn{2}{c}{\textbf{Potential Threat}} & \multicolumn{2}{c}{\textbf{Serious Threat}} & \multicolumn{3}{c}{\textbf{Overall}} \\
\cline{3-11}
 & & Precision & Recall & Precision & Recall & Precision & Recall & Precision & Recall & \textbf{F1 Score} \\
\midrule
\multirow{2}{*}{Qwen2-VL} & No & 54.55 & 65.45 & 25.44 & 38.67 & 67.39 & 28.97 & 49.13 & 44.36& 46.62 \\
 & Yes &  63.43 & 77.27 & 45.76 & 36.00 & 79.80 & 73.83 & 63.00 & 62.37 & \textbf{62.68}\\
\midrule
\multirow{2}{*}{Qwen2.5-VL} & No & 76.32 & 26.36 & 29.19 & 81.33 & 75.56 & 31.78 & 60.35 & 46.49  & 52.52\\
 & Yes & 65.45 & 65.45 & 45.83 & 44.00 & 70.00 & 71.96 & 60.43 & 60.47  & \textbf{60.45}\\
\midrule
\multirow{2}{*}{Llama-3.2-Vision} & No & 42.03 & 26.36 & 21.31 & 52.00 & 62.16 & 21.50 & 41.83 & 33.29  & 37.07\\
 & Yes & 63.64 & 76.36 & 56.10 & 30.67 & 71.43 & 79.44 & 63.72 & 62.16 & \textbf{62.93}\\
\midrule
\multirow{2}{*}{Yi-VL} & No & 36.84 & 6.36 & 25.21 & 81.33 & 100.00 & 1.87 & 54.02 & 29.86 & 38.46 \\
 & Yes & 51.68 & 70.00 & 26.92 & 9.33 & 51.28 & 56.07 & 43.29 & 45.14 & \textbf{44.20}\\
\midrule
\multirow{2}{*}{LLaVA-1.6} & No & 41.18 & 31.82 & 25.25 & 68.00 & 20.00 & 0.93 & 28.81 & 33.58 & 31.01\\
 & Yes & 60.56 & 78.18 & 54.84 & 22.67 & 57.14 & 63.55 & 57.51 & 54.80 & \textbf{56.12}\\
\bottomrule
\end{tabular}}
\end{table*}

\centering
\begin{table*}[htbp]
\caption{Performance Comparison via Individual Fine-Tuning with Type-II Visual Prompt on RailPos Task in Cog-RailSem19 Dataset}
\label{tab:sota_mllm_type2_pos_RailSem19}
\renewcommand{\arraystretch}{1.3}
\resizebox{\textwidth}{!}{ \begin{tabular}{ccccccccccc}
\toprule
\multirow{2}{*}{\textbf{Model}} & \multirow{2}{*}{\textbf{Fine-tuned}} & \multicolumn{2}{c}{\textbf{On Track}} & \multicolumn{2}{c}{\textbf{On Ballast}} & \multicolumn{2}{c}{\textbf{Outside Ballast}} & \multicolumn{3}{c}{\textbf{Overall}} \\
\cline{3-11}
 & & Precision & Recall & Precision & Recall & Precision & Recall & Precision & Recall & \textbf{F1 Score} \\
\midrule
\multirow{2}{*}{Qwen2-VL} & No & 44.04 & 58.54 & 49.33 & 49.33 & 68.52 & 54.81 & 53.96 & 54.23  &54.10 \\
 & Yes & 58.23 & 56.10 & 48.61 & 46.67 & 66.67 & 69.63 & 57.84 & 57.46 & \textbf{57.65}\\
\midrule
\multirow{2}{*}{Qwen2.5-VL} & No & 50.70 & 43.90 & 37.50 & 76.00 & 69.57 & 35.56 & 52.59 & 51.82 & 52.20\\
 & Yes & 66.13 & 50.00 & 53.57 & 60.00 & 64.38 & 69.63 & 61.36 & 59.88 & \textbf{60.61}\\
\midrule
\multirow{2}{*}{Llama-3.2-Vision} & No & 38.10 & 39.02 & 30.86 & 72.00 & 60.61 & 14.81 & 43.19 & 41.95  & 42.56\\
 & Yes & 61.73 & 60.98 & 47.44 & 49.33 & 62.41 & 61.48 & 57.19 & 57.26  & \textbf{57.23}\\
\midrule
\multirow{2}{*}{Yi-VL} & No & 28.65 & 62.20 & 19.47 & 29.33 & 100.00 & 0.74 & 49.37 & 30.76 & 37.90\\
 & Yes & 66.00 & 40.24 & 31.94 & 30.67 & 52.35 & 65.93 & 50.10 & 45.61 & \textbf{47.75}\\
\midrule
\multirow{2}{*}{LLaVA-1.6} & No & 32.76 & 23.17 & 26.92 & 56.00 & 57.14 & 23.70 & 38.94 & 34.29  & 36.47\\
 & Yes & 50.60 & 51.22 & 45.16 & 37.33 & 60.54 & 65.93 & 52.10 & 51.49  & \textbf{51.80}\\
\bottomrule
\end{tabular}}
\end{table*}

\begin{table*}[htbp]
\centering
\caption{Performance Comparison via Individual Fine-Tuning with Type-II Visual Prompt on RailMove Task in Cog-RailSem19 Dataset}
\label{tab:sota_mllm_type2_move_RailSem19}
\renewcommand{\arraystretch}{1.3}
\resizebox{\textwidth}{!}{ \begin{tabular}{ccccccccccc}
\toprule
\multirow{2}{*}{\textbf{Model}} & \multirow{2}{*}{\textbf{Fine-tuned}} & \multicolumn{2}{c}{\textbf{Stationary}} & \multicolumn{2}{c}{\textbf{Threatening Move}} & \multicolumn{2}{c}{\textbf{Non-threatening Move}} & \multicolumn{3}{c}{\textbf{Overall}} \\
\cline{3-11}
 & & Precision & Recall & Precision & Recall & Precision & Recall & Precision & Recall & \textbf{F1 Score} \\
\midrule
\multirow{2}{*}{Qwen2-VL} & No & 53.05 & 79.09 & 28.72 & 36.00 & 70.59 & 22.43 & 50.79 & 45.84 &\textbf{48.19} \\
 & Yes &  53.46 & 77.27 & 29.89 & 34.67 & 65.22 & 28.04 & 49.52 & 46.66 & 48.05\\
\midrule
\multirow{2}{*}{Qwen2.5-VL} & No & 54.47 & 60.91 & 27.01 & 49.33 & 81.25 & 24.30 & 54.24 & 44.85 & \textbf{49.10}\\
 & Yes & 55.46 & 60.00 & 26.32 & 26.67 & 61.86 & 56.07 & 47.88 & 47.58 & 47.73\\
\midrule
\multirow{2}{*}{Llama-3.2-Vision} & No & 43.18 & 51.82 & 22.89 & 25.33 & 46.75 & 33.64 & 37.61 & 36.93  & 37.27\\
 & Yes & 43.98 & 86.36 & 19.23 & 13.33 & 75.00 & 16.82 & 46.07 & 38.84 & \textbf{42.15}\\
\midrule
\multirow{2}{*}{Yi-VL} & No & 44.78 & 27.27 & 24.54 & 70.67 & 100.00 & 0.93 & 56.44 & 32.96 & \textbf{41.61} \\
 & Yes & 53.85 & 6.36 & 25.36 & 93.33 & 66.67 & 1.87 & 48.63 & 33.86  & 39.92\\
\midrule
\multirow{2}{*}{LLaVA-1.6} & No & 40.60 & 49.09 & 31.33 & 34.67 & 26.32 & 18.69 & 32.75 & 34.15 & 33.43\\
 & Yes & 58.12 & 61.82 & 25.18 & 46.67 & 63.89 & 21.50 & 49.06 & 43.33 & \textbf{46.02}\\
\bottomrule
\end{tabular}}
\end{table*}

\begin{table*}[htbp]
\centering
\caption{Performance Comparison via Individual Fine-Tuning with Type-II Visual Prompt on RailThreat Task in Cog-RailSem Dataset}
\label{tab:sota_mllm_type2_threat_RailSem}
\renewcommand{\arraystretch}{1.3}
\resizebox{\textwidth}{!}{ \begin{tabular}{ccccccccccc}
\toprule
\multirow{2}{*}{\textbf{Model}} & \multirow{2}{*}{\textbf{Fine-tuned}} & \multicolumn{2}{c}{\textbf{Safe}} & \multicolumn{2}{c}{\textbf{Potential Threat}} & \multicolumn{2}{c}{\textbf{Serious Threat}} & \multicolumn{3}{c}{\textbf{Overall}} \\
\cline{3-11}
 & & Precision & Recall & Precision & Recall & Precision & Recall & Precision & Recall & \textbf{F1 Score} \\
\midrule
\multirow{2}{*}{Qwen2-VL} & No & 52.27 & 41.82 & 24.84 & 52.00 & 68.09 & 29.91 & 48.40 & 41.24 & 44.53 \\
 & Yes &  65.63 & 76.36 & 58.00 & 38.67 & 75.44 & 80.37 & 66.35 & 65.13 & \textbf{65.74}\\
\midrule
\multirow{2}{*}{Qwen2.5-VL} & No & 74.36 & 26.36 & 28.57 & 90.67 & 86.67 & 12.15 & 63.20 & 43.06   & 51.22\\
 & Yes & 66.38 & 70.00 & 54.93 & 52.00 & 78.10 & 76.64 & 66.47 & 66.21  & \textbf{66.34}\\
\midrule
\multirow{2}{*}{Llama-3.2-Vision} & No & 50.72 & 31.82 & 24.21 & 61.33 & 74.07 & 18.69 & 49.67 & 37.28  & 42.59\\
 & Yes & 67.42 & 80.91 & 68.09 & 42.67 & 74.34 & 78.50 & 69.95 & 67.36 & \textbf{68.63}\\
\midrule
\multirow{2}{*}{Yi-VL} & No & 35.71 & 4.55 & 26.05 & 82.67 & 100.00 & 1.87 & 53.92 & 29.69 & 38.30 \\
 & Yes & 49.70 & 74.55 & 42.86 & 4.00 & 50.00 & 56.07 & 47.52 & 44.87  & \textbf{46.16}\\
\midrule
\multirow{2}{*}{LLaVA-1.6} & No & 51.43 & 32.73 & 25.00 & 70.67 & 50.00 & 3.74 & 42.14 & 35.71 & 38.66\\
 & Yes & 64.71 & 70.00 & 77.27 & 22.67 & 56.29 & 79.44 & 66.09 & 57.37  & \textbf{61.42}\\
\bottomrule
\end{tabular}}
\end{table*}